\documentclass[lettersize,journal]{IEEEtran}
\usepackage{amsmath,amsfonts}
\usepackage{algorithmic}
\usepackage{algorithm}
\usepackage{array}
\usepackage[caption=false,font=normalsize,labelfont=sf,textfont=sf]{subfig}
\usepackage{textcomp}
\usepackage{stfloats}
\usepackage{url}
\usepackage{verbatim}
\usepackage{graphicx}
\usepackage{cite}
\usepackage{graphicx}
\usepackage{booktabs}
\usepackage{color}
\usepackage{colortbl}
\usepackage{xcolor} 
\usepackage{threeparttable}
\usepackage{multirow}
\usepackage{array}
\usepackage{makecell}
\newcolumntype{C}[1]{>{\centering}p{#1}}
\usepackage{epsfig}
\usepackage{float}
\usepackage{bbding} 
\usepackage{eccvabbrv}
\usepackage{subfig}

\definecolor{Gray}{gray}{0.9}
\definecolor{baselinecolor}{gray}{0.9}
\begin{document}

\title{A Hierarchical Semantic Distillation Framework for Open-Vocabulary Object Detection}

\author{Shenghao Fu, Junkai Yan, Qize Yang, Xihan Wei, Xiaohua Xie*, Wei-Shi Zheng*
\thanks{S. Fu and J. Yan are with the School of Computer Science and Engineering, Sun Yat-sen University, Guangzhou 510006, China and Peng Cheng Laboratory, China. Q. Yang and X. Wei are with the Tongyi Lab, Alibaba Group, China. X. Xie and W.-S. Zheng are with the School of Computer Science and Engineering, Sun Yat-sen University, Guangzhou, Guangdong, China, also with the Guangdong Key Laboratory of Information Security Technology, Sun Yat-sen University, Guangzhou, Guangdong, China, and also with the Key Laboratory of Machine Intelligence and Advanced Computing, Sun Yat-sen University, Ministry of Education, Guangzhou, Guangdong, China. 

Corresponding author: Xiaohua Xie and Wei-Shi Zheng.}}


%


\maketitle

\begin{abstract}
  Open-vocabulary object detection (OVD) aims to detect objects beyond the training annotations, where detectors are usually aligned to a pre-trained vision-language model, \eg, CLIP, to inherit its generalizable recognition ability so that detectors can recognize new or novel objects.
  However, previous works directly align the feature space with CLIP and fail to learn the semantic knowledge effectively.
  In this work, we propose a hierarchical semantic distillation framework named HD-OVD to construct a comprehensive distillation process, which exploits generalizable knowledge from the CLIP model in three aspects.
  In the first hierarchy of HD-OVD, the detector learns fine-grained \textit{instance-wise semantics} from the CLIP image encoder by modeling relations among single objects in the visual space. 
  Besides, we introduce text space novel-class-aware classification to help the detector assimilate the highly generalizable \textit{class-wise semantics} from the CLIP text encoder, representing the second hierarchy.
  Lastly, abundant \textit{image-wise semantics} containing multi-object and their contexts are also distilled by an image-wise contrastive distillation.
  Benefiting from the elaborated semantic distillation in triple hierarchies, our HD-OVD inherits generalizable recognition ability from CLIP in instance, class, and image levels. Thus, we boost the novel AP on the OV-COCO dataset to 46.4\% with a ResNet50 backbone, which outperforms others by a clear margin. We also conduct extensive ablation studies to analyze how each component works. Code is available at \url{https://github.com/iSEE-Laboratory/HD-OVD}.
\end{abstract}


\section{Introduction}
\label{sec:intro}

\begin{figure}[tb]
  \centering
  \includegraphics[width=0.9\linewidth]{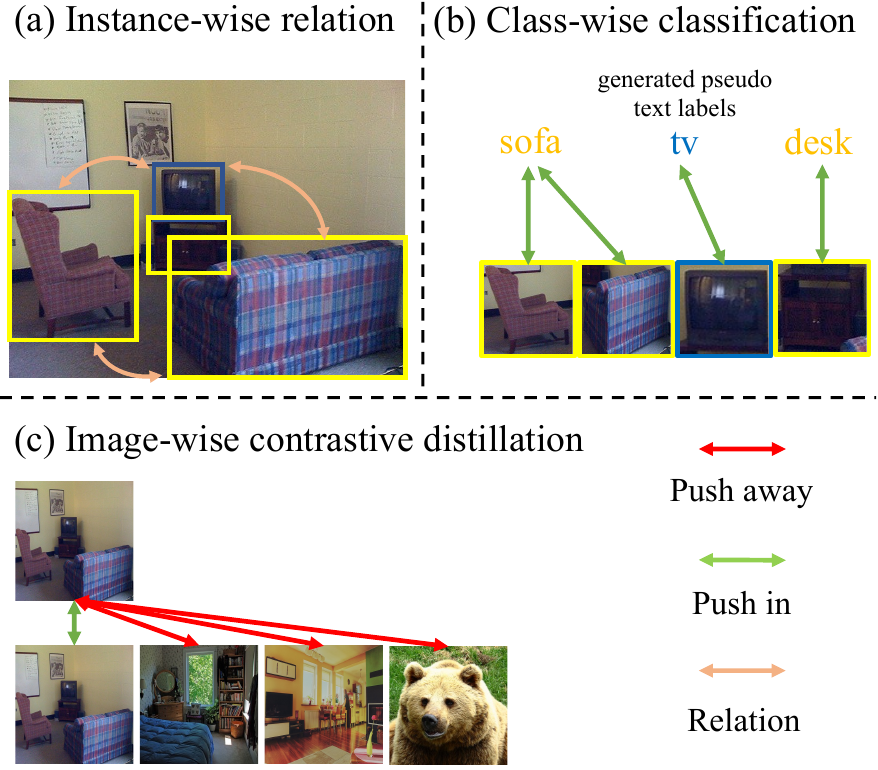}
  \caption{The illustration of three-level semantic knowledge modeling in HD-OVD. It consists of (a) instance-wise relation modeling, (b) class-wise novel-class-aware classification, and (c) image-wise contrastive distillation.
  Distilling hierarchical semantics from CLIP equips HD-OVD with strong open-vocabulary ability. Base and novel class boxes are colored in \textcolor{yellow}{yellow} and \textcolor{blue}{blue}. Best viewed in color.}
  \label{fig:intro}
\end{figure}

\begin{table*}[t]
  \renewcommand{\arraystretch}{1.2}
  \caption{Classification error analysis on OV-COCO validation dataset. `\#Recall' and `\#Selected' denote the number of novel objects recalled by a total of 300 object queries (or 1,000 RoIs) and selected as the final 100 predictions, respectively. `\#Base', `\#Novel', and `\#Correct' are the number of selected object queries whose max classification score belongs to base classes, inaccurate novel classes, and correct novel classes, respectively. There are a total of 4,582 novel objects in the dataset. Although the model training with only base class has a high `\#recall', only partial object queries covering novel objects are selected due to their low classification scores. Further, most selected queries are misclassified as base classes. Our method has solved the problem effectively.}
  \centering
  {\begin{tabular}{l|ccc|cc|ccc}
    \hline
    Methods  &  AP50$^{\text{box}}$ & AP50$^{\text{box}}_{\text{base}}$ & AP$^{\text{box}}_{\text{novel}}$ & \#Recall & \#Selected & \#Base & \#Novel & \#Correct \\
    \hline
    Base-only & 46.3 & 61.8 & 2.8 & 3681 & 2334 & 2320 & 5 & 9 \\
     \hline
    OADP~\cite{oadp} & 48.8 & 55.0 & 31.3 & 3906 & 3754 & 2331 & 98 & 1325 \\
     \hline
    HD-OVD (Ours) & 57.4 & 61.3 & 46.4 & 3930 & 3550 & 876 & 159 & 2515 \\
    \hline
  \end{tabular}}
  \label{tab:error_analysis}
  \vspace{-0.5em}
\end{table*}

\begin{figure*}[tb]
  \centering
  \includegraphics[width=0.7\linewidth]{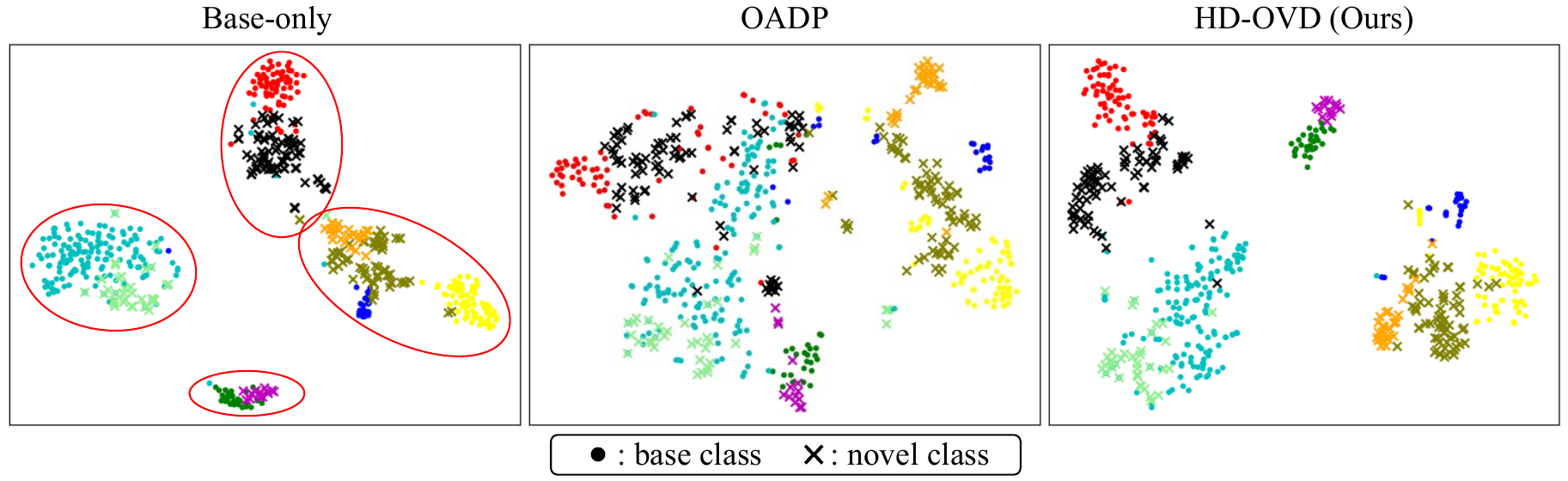}
  \vspace{-0.5em}
  \caption{t-SNE plots of RoI features for each model. Different colors represent different classes. The model trained with base classes struggles to separate similar base and novel classes, as shown in red circles. Our HD-OVD can separate them clearly.}
  \label{fig:tsne}
  \vspace{-0.5em}
\end{figure*}

\IEEEPARstart{O}{bject} detection is a fundamental computer vision task and has made significant progress in the era of deep learning. 
Although detectors~\cite{ren2015faster, tian2019fcos, detr, deformabledetr, fu2023asag, 9416174, 9684715, mo2024bridge, fu2024frozen} perform well under the closed-set validation, they cannot generalize to novel classes that are unlabeled in the training set. As shown in Table~\ref{tab:error_analysis}, training with only base class annotations, the detector~\cite{adamixer} fails to generalize to novel classes (2.8\% novel AP). As a na\"{i}ve solution, expanding the number of labeled classes and retraining the detector may work, but it greatly limits the practical usage. Thus, open-vocabulary object detection (OVD)~\cite{wu2024towards, ovr-cnn, vild} aiming to detect labeled base classes and unlabeled novel classes simultaneously has caught progressive attention.

Although detectors trained with only base classes cannot correctly classify objects of novel classes, we find that such detectors can localize most of them accurately. Taking a query-based detector~\cite{adamixer} as an example, it can localize more than 80\% of novel-class objects (3,681), as shown in the first row of Table~\ref{tab:error_analysis}. However, only about 63\% of the correctly localized objects (2,334) are selected as the final predictions since the rest have low classification scores and are thus misclassified as background and filtered out. Furthermore, the highest classification scores of these 2,334 selected boxes covering novel objects nearly always fall to base classes. This phenomenon indicates that learning highly distinguishable semantic knowledge for novel classes is one of the keys to OVD.\looseness=-1

To learn distinguishable semantics for promoting recognition ability, recent works have exploited the transferable knowledge on large vision-language models~\cite{clip} pre-trained with large-scale image-text pairs.
Motivated by the impressive generalization ability of CLIP~\cite{clip}, most recent methods propose to obtain the open-vocabulary ability by (1) aligning the Region-of-Interest (RoI) or query features with the CLIP image encoder~\cite{vild, oadp, BARON, dk-detr} or (2) directly matching the objects with the class embeddings output by the CLIP text encoder using paired image-text data such as image captions~\cite{vldet, goat} or image-level labels~\cite{detic, feng2022promptdet}.
The methods previously mentioned explore the semantic knowledge within CLIP from individual perspectives and fail to fully inherit the rich hierarchical knowledge that CLIP encodes at the instance, class, and image levels.
As shown in the second row of Tabel~\ref{tab:error_analysis}, OADP~\cite{oadp}, a state-of-the-art method that strongly aligns the feature space with CLIP image encoder, still suffers from misclassifying novel-class objects as base classes. On the other hand, Figure~\ref{fig:tsne} illustrates the RoI features of OADP for each class, also suggesting that the novel classes can hardly be separated from base classes.

In this work, we propose a hierarchical semantic distillation framework to inherit abundant distinguishable semantic knowledge from CLIP.
We explore the semantic knowledge from three aspects hierarchically, as shown in Figure~\ref{fig:intro}: 
(1) From the \textbf{instance-wise} aspect, we explicitly model the novel-and-base and novel-and-background relationships by selecting some class-agnostic pseudo boxes out of the background and distinguishing them from base classes. The distinguishable and fine-grained features of individual objects from the CLIP image encoder help the detector correctly separate novel classes from base classes and background.
(2) From the \textbf{class-wise} aspect, we design an effective pseudo-labeling pipeline to assign a generated pseudo text label for each class-agnostic pseudo box to constrain the classification of the box, avoiding the box being classified as background. Such supervision enables the distillation of highly generalizable category knowledge from the CLIP text encoder to the detector and enhances its class-wise awareness.
(3) Considering a scene usually comprises multiple objects of various classes and surrounding context, we also incorporate an \textbf{image-wise} distillation. We transfer this structural and context information from the CLIP image encoder to the detector through an image-wise contrastive distillation technique.
Through the above three-level distillation, the detector inherits hierarchical semantic knowledge from CLIP, forming a comprehensive knowledge transfer.

Extensive experiments on OV-COCO and OV-LVIS benchmarks validate the superiority of our HD-OVD learning framework. Especially, in the challenging OV-COCO dataset that has only 48 base classes and is hard to learn transferable knowledge, our method outperforms previous state-of-the-art methods by 5\% to 10\% AP$_{novel}$ without leaking any information of novel classes. As shown in the last row of Table~\ref{tab:error_analysis} and Figure~\ref{fig:tsne}, our HD-OVD can classify novel classes correctly beyond base classes and background. 

We summarize our contributions as follows:
\begin{itemize}
    \item \textbf{Integral and complementary distillation.} We propose to distill novel class knowledge by modeling rich details within each instance, the commonly shared features within a class, and context information within a scene, forming a semantic hierarchy. 
    \item \textbf{Effective pseudo-labeling pipeline.} We propose a novel pseudo text label generation pipeline for class-agnostic pseudo boxes, which is the first one to boost the discriminability of novel classes by a frozen caption model.
    \item \textbf{High open-vocabulary performance.} With the integral distillation pipeline and novel pseudo-labeling mechanism, we can model base and novel classes in a unified space, which can better distinguish novel classes from base classes and background, achieving high performance on various datasets.
\end{itemize}

\section{Related Works}

\begin{figure*}[tb]
  \centering
  \includegraphics[width=0.81\linewidth]{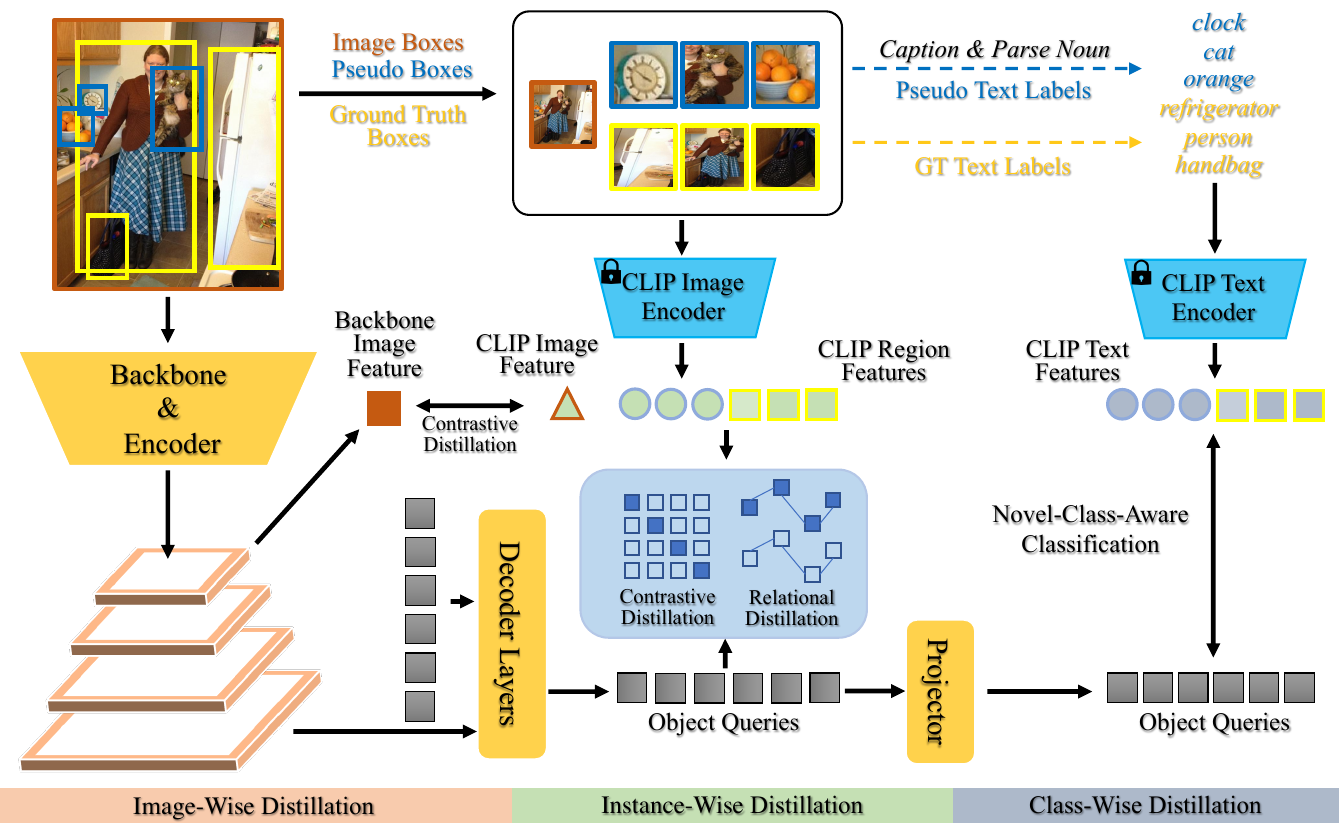}
  \vspace{-0.5em}
  \caption{The overview of HD-OVD, which distills semantics from CLIP to a query-based detector hierarchically. 
  At the image level, the backbone feature of the image is aligned with the CLIP global image feature.
  At the instance level, relations among various single instances from the CLIP image encoder are transferred to the detector. The class-level distillation conducts a novel-class-aware classification to enable the detector to inherit the high-level semantics from the CLIP text encoder.
  To save training computation cost, all CLIP features are pre-extracted in an offline manner. Best viewed in color.}
  \label{fig:model_overview}
  \vspace{-0.5em}
\end{figure*}

\subsection{Traditional Object Detection}
Traditional object detectors can be divided into one-stage \cite{lin2017feature, lin2017focal, tian2019fcos} and two-stage~\cite{ren2015faster} methods based on whether using a Region Proposal Network~\cite{ren2015faster} to extract proposals. Recently, with the arising of vision transformers~\cite{dosovitskiy2020image,  10041780,10380775}, query-based detectors~\cite{detr, deformabledetr, adamixer} utilizing transformer encoder-decoder architecture to reason objects globally has achieved an elegant framework and higher performance. Despite significant structural differences, all these detectors are trained and tested on a predefined category set. Before calculating losses, each prediction will be assigned to one of the base class ground truths or background based on the prediction score and localization quality. In the closed-world setting, bounding boxes that cover novel objects will be assigned to the background, thus limiting open-vocabulary ability.

\subsection{Open-Vocabulary Object Detection}

Open-vocabulary detectors can detect arbitrary classes despite only training with base class annotations. Recent works can be categorized into three parts: knowledge-distillation-based methods~\cite{vild, ma2022open,oadp,BARON,dk-detr,shi2023edadet}, pseudo-labeling-based methods~\cite{detic, feng2022promptdet, gao2022open, vldet, ma2023codet, zhao2022exploiting, pham2024lp, huynh2022open, zhao2024taming, xu2023dstdet} and prompt-engineering-based methods~\cite{detpro, feng2022promptdet, ov-detr, kim2024retrieval, zhao2024scene}.

Motivated by the impressive zero-shot ability of large vision-language models (\eg CLIP~\cite{clip}), distilling knowledge from CLIP has been a widely explored direction for addressing OVD. ViLD~\cite{vild} directly aligns RoI features with CLIP features with L1 loss. However, CLIP is pretrained with image-text contrastive learning and focuses on image-level features and main objects in the image, while detectors pursue fine-grained region-level representations. Such a feature representation dilemma makes direct feature alignment less effective. OADP~\cite{oadp} distills the feature from multiple box sizes. BARON~\cite{BARON} explores aligning multiple objects as a whole rather than isolated individual objects. EdaDet~\cite{shi2023edadet} notices that novel classes are easily misclassified as similar base classes and pursues a fine-grained distillation using dense alignment. DK-DETR~\cite{dk-detr} explores the relationships within potential novel classes. However, previous works only explore a single aspect of distillation, making the knowledge transfer less effective. In this work, to pursue an integral and complementary distillation, we explore distillation in a semantic hierarchy.

Pseudo-labeling-based methods explore weak supervision from pseudo labels. Some methods match the image-level labels or noun phrases in captions to proposals with some heuristic rules, \eg max activation~\cite{gao2022open}, max size~\cite{detic}, bipartite matching~\cite{vldet}, co-occurrence~\cite{ma2023codet}. VL-PLM~\cite{zhao2022exploiting} and LP-OVOD~\cite{pham2024lp} directly use CLIP to classify top-scored proposals into predefined categories. Some methods~\cite{huynh2022open, feng2022promptdet, zhao2024taming, xu2023dstdet} also use a self-trained detector or segmentor as the labeler. The above methods are all \textbf{matching-based methods}, in which a predefined label set is used for matching. Such a dictionary largely restricts the diversity and accuracy of the generated pseudo labels. On the contrary, our HD-OVD utilizes a \textbf{generation-based method} to obtain pseudo labels by using a caption model to generate instance-specific captions as pseudo labels, which provides richer information for the specific instance and is scalable with the caption model.

Another line of research explores diverse object descriptions to build an object-aware visual-language space~\cite{glip, grounding_dino, 10480273, fu2025llmdet}. These works use region-word contrastive learning to learn discriminative features, achieving promising performance in the wild. These methods are highly based on massive grounding data, requiring intensive human annotations.

\section{A Hierarchical Semantic Distillation Framework}

\subsection{Overview}

In open-vocabulary object detection, the detector has trained with base class $C_{base}$ annotations and is expected to detect arbitrary classes $C_{test}$ prompted by the class name during testing, \eg \emph{`superman'}, where $C_{test} = C_{base} \cup C_{novel}$, $C_{base} \cap C_{novel} = \emptyset$, and $C_{novel} \neq \emptyset$.

In this work, we aim to equip a query-based detector with open-vocabulary ability. We propose to learn novel classes using \textbf{Hierarchical Semantic Distillation} from fine-grained instance-wise, high-semantic class-wise, and comprehensive image-wise.
Figure~\ref{fig:model_overview} illustrates our overall framework.

Since annotations only include base classes and lack information about novel classes, using only base class annotations is insufficient to inherit abundant semantics from CLIP. 
As a result, we select several high-confidence proposals that do not overlap with base classes out of the background and treat them as pseudo boxes to partially compensate for the missing novel boxes. Then the \textbf{instance-wise} relationships among base and pseudo boxes are modeled by distilling the detailed instance-level knowledge from the CLIP image encoder (Section~\ref{image-distill}), forming the first hierarchical distillation item.

Except for instance-wise modeling, we further propose a \textbf{class-wise} distillation to transfer the high-semantic class-level knowledge from CLIP, which is regarded as the second hierarchy. 
We construct a pseudo text label generation pipeline to complete the missing class labels for the pseudo boxes mentioned before.
Incorporating the complemented bounding boxes and class labels of potential novel classes, along with ground truth labels of base classes, we can align the features across different classes to the CLIP text encoder, enhancing their distinguishability (Section~\ref{text-distill}).

In addition, the last hierarchy is conducted \textbf{image-wise} via an image-level contrastive distillation, where the knowledge among multi-instance, multi-class, and the related context is distilled from the CLIP image encoder, facilitating a comprehensive understanding of the image (Section~\ref{global-distill}).

Together, the above three distillation hierarchies compose our entire HD-OVD learning framework. Through our HD-OVD, the detector is able to distinguish each individual object in a fine instance-level granularity, absorbs the class-wise generalizable knowledge, and inherits the image understanding ability from the CLIP teacher, thus achieving a better open-vocabulary capability.

\begin{figure*}[tb]
  \centering
  \includegraphics[width=0.9\linewidth]{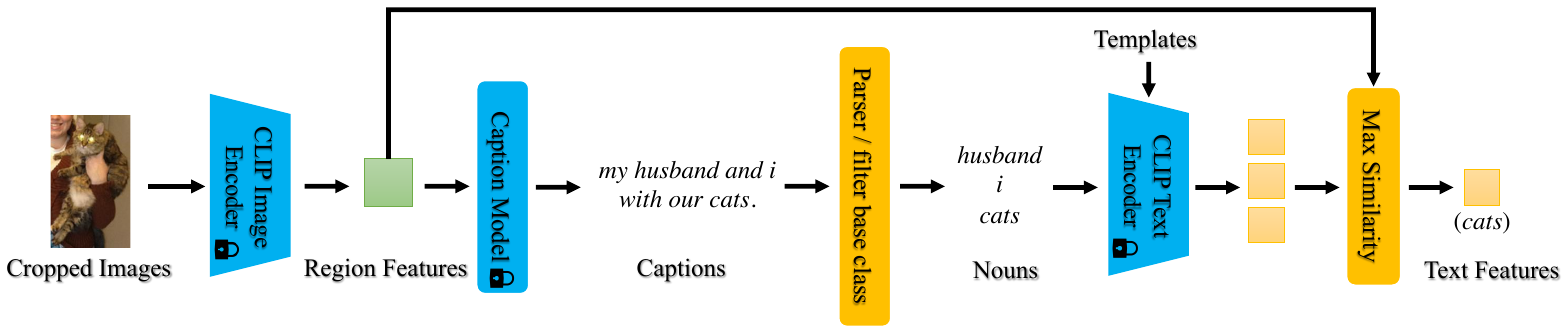}
  \vspace{-0.5em}
  \caption{The pipeline for generating pseudo text labels. We first generate captions for each region based on CLIP region features. Then, nouns are extracted by a grammar parser. The noun having a max CLIP similarity with the region features is selected as the final pseudo text label.}
  \label{fig:preprocess}
  \vspace{-0.5em}
\end{figure*}

\subsection{Image-Space Instance-Wise Distillation}
\label{image-distill}

In our Hierarchical Semantic Distillation, we begin by aligning object queries with CLIP region features, which are rich in detail for each object. This approach preserves the distinctiveness of each object, resulting in a more fine-grained distillation. Notably, CLIP region features encompass both base-class objects and pseudo-labeled objects potentially containing novel objects.
To obtain the pseudo boxes for potential novel objects, we first collect several proposals with high confidence using an off-the-shelf class-agnostic proposal network. Proposals overlapping with base class annotations beyond a predefined threshold are discarded. The remaining proposals are regarded as pseudo boxes, denoted as $\mathbf{B}_{pse}$. Similar to base class boxes from ground-truth (GT) $\mathbf{B}_{gt}$, pseudo boxes are assigned to object queries using the default label assignment strategy, \ie, bipartite matching~\cite{detr}. With all bounding boxes $\mathbf{B}=\mathbf{B}_{gt} \cup \mathbf{B}_{pse}$, CLIP region features $\mathbf{E}^V$ can be obtained by feeding the cropped images into the CLIP image encoder $\mathcal{V}(\cdot)$, formulated as $\mathbf{E}^V = \mathcal{V}(\text{Crop}(I,\mathbf{B})) = \{e^v_i\}_{i=1}^{n}$, where $n$ denotes the number of bounding boxes $\mathbf{B}$ (GT and pseudo boxes) and Crop represents the operation of cropping a region from the image.

The instance-wise distillation is carried out between the matched object queries and the ground truth together with pseudo boxes. Specifically, the matched object queries $\mathbf{Q} = \{q_i\}_{i=1}^n$ are first projected to the same dimension as CLIP region features $\mathbf{E}^V \in \mathbb{R}^{n \times d}$. Then, we employ contrastive and relational distillation techniques \cite{dk-detr} for knowledge transfer.
Contrastive knowledge distillation (CKD) aims to minimize the distances between positive pairs while pushing away negative pairs. In our case, object queries and their corresponding CLIP region embeddings serve as positive pairs, while others are negative samples. Cosine similarity is used as the distance function, and binary cross-entropy loss is used, formulated as:
\begin{align}
  \mathcal{L}_{ckd}^{ins} & = -\frac{1}{n} \sum_{i=1}^{n} \left( \log{(p_{i,i}^{dis})} + \sum_{j=1, j \neq i}^{n}\log{(1-p_{i,j}^{dis})} \right), \\
  p_{i,j}^{dis} & = \text{sigmoid}(\cos{(q_i, e^v_j)} \cdot \tau_{ckd}),
\end{align}
where $\tau_{ckd}$ is the temperature coefficient of CKD. To enlarge the number of negative samples, we further gather CLIP region embeddings from all images in the same GPU and pad additional CLIP region embeddings from a memory queue~\cite{moco}. Moreover, relational knowledge distillation (RKD) aims to model the relationship between individual objects such that the distillation can be performed in a continuous space. We first compute the relational matrices $\mathbf{R}_q^V = \mathbf{Q}\mathbf{Q}^{\mathsf{T}} \in \mathbb{R}^{n \times n}$ and $\mathbf{R}_e^V = (\mathbf{E}^V)(\mathbf{E}^V)^{\mathsf{T}} \in \mathbb{R}^{n \times n}$ for object queries and CLIP region features separately. Cosine similarity is once again used as the relation metric. We follow DK-DETR~\cite{dk-detr} using Kullback-Leibler divergence as the loss function,
\begin{equation}
  \mathcal{L}_{rkd}^{ins} = -\frac{1}{n} \sum_{i=1}^{n}KL\left( \text{softmax}(\frac{\mathbf{R}_q^V[i,:]}{1/\tau_{rkd}}) || \text{softmax}(\frac{\mathbf{R}_e^V[i,:]}{1/\tau_{rkd}}) \right),
\end{equation}
where $\tau_{rkd}$ is the temperature coefficient of RKD. Different from DK-DETR, we do not use an auxiliary branch and instead perform distillation on both base and novel-class objects in a unified way, which not only preserves the performance of base classes but also explicitly separates novel-class objects from base-class objects, promoting novel classes to be more distinguishable.

\subsection{Text-Space Class-Wise Distillation}
\label{text-distill}
Based on image-space instance-wise distillation, we further distill class-wise knowledge from the CLIP text encoder by aligning object queries with corresponding class embeddings since language is a more general high-level semantic compared with visual cues, which can maintain consistency within classes. Although text labels for base-class objects $\mathbf{T}_{gt}$ are available, the text labels for pseudo boxes are missing. To generate pseudo text labels, we use an off-the-shelf caption model to generate captions for each region adaptively, as shown in Figure~\ref{fig:preprocess}.
However, the generated captions are not optimal. First, the captions are free-form texts which are inconsistent with base class labels. The captions may contain some unrelated, distracting, or even hallucinatory words. Moreover, the caption model may fail to generate accurate captions due to limited capacity or inaccurate pseudo boxes. All these problems will confuse the detector, resulting in poor performance. To ensure that the generated captions are consistent in format with base class labels, we use a grammar parser~\cite{nltk} to extract the noun phases from the captions and then select the noun phase with a max CLIP similarity corresponding to the region feature as the final pseudo text label $\mathbf{T}_{pse}$. We use the CLIP matching scores $s_j$ as the correctness scores since they describe the degree of matching between nouns and objects. The scores also apply to the classification loss to reduce the negative impact of inaccurate labels. Text labels are also encoded to the embedding format $\mathbf{E}^T = \mathcal{T}(\mathbf{T}_{gt} \cup \mathbf{T}_{pse}) = \{e^t\}$.

With pseudo boxes paired with text labels, the matched object queries are aligned with class embeddings generated by the CLIP text encoder using novel-class-aware classification loss, in which novel-class objects will be assigned to their corresponding text labels instead of the background. 
Specifically, we first replace the classifier with base class embeddings $E_{base}^T$ following common practices and expand it with pseudo text embeddings $E_{pse}^T$, making pseudo novel objects distinct from base classes and background. Thus, the final classifier in training phase is $W_{cls}^{train} = \{ e_{{base}_1}^t, ..., e_{{base}_M}^t, e_{{pse}_1}^t, ..., e_{{pse}_K}^t\}$, where $M$, $K$ denote the number of base and pseudo class embeddings. Further, a weighting mechanism is used to reduce the negative impacts of noisy pseudo-text labels. The classification loss is carried out by:
\begin{align}
  &\mathcal{L}_{cls} = -\frac{1}{N}\sum_{i=1}^{N}\sum_{j=1}^{M+K} \alpha_j y_{i,j}Focal(p_{i,j}^{cls}), \\
  \alpha_j = & \begin{cases}
        1, \, & \text{if } 1 \leq j \leq M \text{ (base class)},\\
        \sigma(s_j), \, & \text{if } M+1 \leq j \leq M+K \text{ (pseudo class)}\\
  \end{cases}
\end{align}
where $p_{i,j}^{cls} = \text{sigmoid}(\cos{(\hat{q}_i, e^t_j)} \cdot \tau_{cls})$ is the classification score between the $i$-th object query and the $j$-th class embedding, $\sigma$ is sigmoid function, and $s_j$ is the CLIP matching score between $j$-th class embedding and its region embedding after standardization. $N$ denotes the number of object queries. We separate the distillation space by projecting object queries to text space $\hat{\mathbf{Q}} = \text{MLP}(\mathbf{Q}) = \{\hat{q}\}$ to avoid conflicting learning. Following standard query-based detectors~\cite{detr, deformabledetr, adamixer}, $Focal$ loss~\cite{lin2017focal} is used as the classification loss. 

\subsection{Image-Wise Global Distillation}
\label{global-distill}
In addition to semantics for individual objects in image space and class-level semantics in text space, the whole image with various objects of multiple classes and diverse environments also contains abundant information beyond an individual object and a single class. Recent works~\cite{BARON, fang2023simple} show that CLIP can implicitly capture the compositional structure of semantic concepts in a scene, which is helpful to open-vocabulary recognition. Inspired by the global semantics in CLIP, we propose to perform image-level contrastive distillation to align detector's backbone features with CLIP image features in a global manner. 

Specifically, in addition to object boxes, an additional image box $\mathbf{B}_{G}$ is used to extract CLIP global feature $\mathbf{E}^G = \mathcal{V}(\text{Crop}(I,\mathbf{B}_{G})) = \{e^g\}$. The global image feature $x^g \in \mathbb{R}^{C}$ from backbone is the averaged feature from the feature pyramid $\{X_i\}_{i=0}^{l-1} \in \mathbb{R}^{C\times H_i \times W_i}$:
\begin{align}
  X &= \frac{1}{l} \sum_{i=0}^{l-1} \text{interpolate}(X_i), \\
  x^g &= \text{AvgPool}(X),
\end{align}
where we should interpolate different feature maps to the same spatial size. After projecting the backbone features to the same dimension as CLIP features, the standard bi-directional contrastive distillation can be performed between them:
\begin{align}
  \mathcal{L}_{ckd}^{img} & = -\frac{1}{2} \sum_{i=1}^{m} \left( \log{(p_{x,e}^i)} + \log{(p_{e,x}^i)} \right), \\
  \log{(p_{x,e}^i)} & = \frac{\exp{(\cos{(x^g_i, e^g_i)} \cdot \tau_{ckd})}}{\sum_{j=1}^{m}\exp{(\cos{(x^g_i, e^g_j)} \cdot \tau_{ckd})}}, \\
  \log{(p_{e,x}^i)} & = \frac{\exp{(\cos{(x^g_i, e^g_i)} \cdot \tau_{ckd})}}{\sum_{j=1}^{m}\exp{(\cos{(x^g_j, e^g_i)} \cdot \tau_{ckd})}}.
\end{align}
where $m$ is the number of images. Similar to instance-wise contrastive distillation, we pad some negative embeddings from a memory queue.

\begin{table*}[t]
  \caption{Open-vocabulary object detection results on OV-COCO. `Detector' means the base detector used by each method. `Strict' means whether novel class names are known during the training. For the training time, 1$\times$ schedule denotes 12 epochs and we report the approximate number of schedules in integer format. AP$_{n}$ is the main metric and is highlighted.}
  \centering 
      \begin{tabular}{l|c|c|c|c|c|ccc}
        \hline
        \multirow{2}{*}{Methods} & \multirow{2}{*}{Backbone} & \multirow{2}{*}{Detector} & External & Training & \multirow{2}{*}{Strict} & \multicolumn{3}{c}{OV-COCO}\\
        & & & Dataset & Time & & AP & AP$_{b}$ & AP$_{n}$ \\
        \hline
        PromptDet~\cite{feng2022promptdet} & R50 & Mask-RCNN & LAION-Novel & 2$\times$ & \XSolidBrush & \textcolor{gray}{50.6} & - & 26.6 \\
        ViLD-Ens.~\cite{vild} & R50 & Mask-RCNN & - & 2$\times$ & \checkmark & \textcolor{gray}{51.4} & \textcolor{gray}{59.7} & 27.7  \\
        Detic~\cite{detic} & R50 & Faster-RCNN & COCO Caption & 1$\times$ & \XSolidBrush & \textcolor{gray}{45.0} & \textcolor{gray}{47.1} & 27.8 \\
        F-VLM~\cite{fvlm} & R50 & Mask-RCNN & -  & 2$\times$ & \checkmark & \textcolor{gray}{39.6} & - & 28.0  \\
        CondHead~\cite{wang2023learning} & R50 & Mask-RCNN & - & 2$\times$ & \checkmark & \textcolor{gray}{52.7} & \textcolor{gray}{60.8} & 29.8  \\
        OADP~\cite{oadp}  & R50 & Mask-RCNN & - & 1$\times$ & \checkmark & \textcolor{gray}{47.2} & \textcolor{gray}{53.3} & 30.0  \\
        ProxyDet~\cite{jeong2024proxydet} & R50 & Faster-RCNN & COCO Caption & 1$\times$ & \checkmark & \textcolor{gray}{46.8} & \textcolor{gray}{52.6} & 30.4 \\
        CoDet~\cite{ma2023codet} & R50 & Faster-RCNN & COCO Caption & 1$\times$ & \checkmark & \textcolor{gray}{46.6} & \textcolor{gray}{52.3} & 30.6 \\
        GOAT~\cite{goat} & R50 & Faster-RCNN & COCO Caption & 1$\times$ & \checkmark & \textcolor{gray}{46.1} & \textcolor{gray}{51.3} & 31.7 \\
        VLDet~\cite{vldet} & R50 & Faster-RCNN & COCO Caption & 1$\times$ & \checkmark & \textcolor{gray}{45.8} & \textcolor{gray}{50.6} & 32.0 \\
        RALF~\cite{kim2024retrieval} & R50 & Mask-RCNN & - & - & \checkmark & \textcolor{gray}{49.0} & \textcolor{gray}{54.5} & 33.4 \\
        BARON~\cite{BARON}  & R50 & Faster-RCNN & - & 2$\times$ & \checkmark & \textcolor{gray}{53.5} & \textcolor{gray}{60.4} & 34.0  \\
        BARON+~\cite{BARON} & R50 & Faster-RCNN & COCO Caption & 2$\times$ & \XSolidBrush & \textcolor{gray}{51.7} & \textcolor{gray}{54.9} & 42.7 \\
        SAMP~\cite{zhao2024scene} & R50 & Faster-RCNN & - & 1$\times$ & \checkmark & \textcolor{gray}{54.2} & \textcolor{gray}{61.4} & 34.8 \\
        DVDet~\cite{jin2024llms}  & R50 & Faster-RCNN & - & 1$\times$ & \checkmark & \textcolor{gray}{35.8} & \textcolor{gray}{57.0} & 35.8  \\
        object-centric ovd~\cite{object-centric-ovd} & R50 & Faster-RCNN & COCO Caption & 3$\times$ & \XSolidBrush & \textcolor{gray}{49.4} & \textcolor{gray}{54.0} & 36.6 \\
        SAS-Det~\cite{zhao2024taming} & R50 & Faster-RCNN & - & 1$\times$ & \checkmark & \textcolor{gray}{53.0} & \textcolor{gray}{58.5} & 37.4 \\
        LP-OVOD~\cite{pham2024lp} & R50 & Faster-RCNN & - & 2$\times$ & \checkmark & \textcolor{gray}{55.2} & \textcolor{gray}{60.5} & 40.5 \\
        \rowcolor{Gray} HD-OVD (Ours) & R50 & Faster-RCNN & - & 1$\times$ & \checkmark & \textcolor{gray}{52.9} & \textcolor{gray}{55.7} & \textbf{45.2} \\
        \rowcolor{Gray} HD-OVD (Ours) & R50 & Faster-RCNN & - & 2$\times$ & \checkmark & \textcolor{gray}{54.7} & \textcolor{gray}{57.7} & \textbf{46.3} \\
        \hline
        OV-DETR~\cite{ov-detr} & R50 & Deformable DETR & - & 4$\times$ & \XSolidBrush & \textcolor{gray}{52.7} & \textcolor{gray}{61.0} & 29.4\\
        DK-DETR~\cite{dk-detr} & R50 & Deformable DETR & - & 6$\times$ & \checkmark & - & \textcolor{gray}{61.1} & 32.3 \\
        CORA~\cite{wu2023cora} & R50 & DAB-DETR & - & 3$\times$ & \checkmark & \textcolor{gray}{35.4} & \textcolor{gray}{35.5} & 35.1 \\
        EdaDet~\cite{shi2023edadet} & R50 & Deformable DETR & - & 4$\times$ & \checkmark & \textcolor{gray}{52.5} & \textcolor{gray}{57.7} & 37.8 \\
        \rowcolor{Gray} HD-OVD (Ours) & R50 & AdaMixer & - & 1$\times$ & \checkmark & \textcolor{gray}{57.4} & \textcolor{gray}{61.3} & \textbf{46.4} \\
        \hline
        RegionCLIP~\cite{zhong2022regionclip} & R50x4 & Faster RCNN & CC3M & 1$\times$ & \checkmark & \textcolor{gray}{55.7} & \textcolor{gray}{61.6} & 39.3 \\
        BIND~\cite{zhang2024exploring} & ViT-B & DETR & - & 8$\times$ & \checkmark & \textcolor{gray}{50.2} & \textcolor{gray}{54.7} & 36.3 \\
        BIND~\cite{zhang2024exploring} & ViT-L & DETR & - & 8$\times$ & \checkmark & \textcolor{gray}{54.8} & \textcolor{gray}{58.3} & 41.5 \\
        CORA~\cite{wu2023cora} & R50x4 & DAB-DETR & - & 3$\times$ & \checkmark & \textcolor{gray}{43.8} & \textcolor{gray}{44.5} & 41.7 \\
        CORA~\cite{wu2023cora} & R50x4 & DAB-DETR & COCO Caption & 3$\times$ & \checkmark & \textcolor{gray}{56.2} & \textcolor{gray}{60.9} & 43.1 \\
        CLIPSelf~\cite{wu2023clipself} & ViT-L & Mask-RCNN & - & - & \checkmark & \textcolor{gray}{-} & \textcolor{gray}{-} & 44.3 \\
        \rowcolor{Gray} HD-OVD (Ours) & Swin-B & AdaMixer & - & 1$\times$ & \checkmark & \textcolor{gray}{62.7} & \textcolor{gray}{66.3} & \textbf{52.7} \\
        \hline
      \end{tabular}
  \label{tab:main_result_coco}
\end{table*}

\begin{table*}[ht]
  \caption{Mask and box results on the OV-LVIS dataset. All models use ResNet-50 as the backbone. `DETR' means using a DETR-like detector. `Strict' means whether novel class names are known during the training. IN-L~\cite{deng2009imagenet}: a subset from ImageNet containing 997 classes overlapping with LVIS. LAION-Novel~\cite{schuhmann2021laion}: a subset from LAION-400M containing 300 images for each novel class. CC3M~\cite{cc3m}: a large-scale image-text dataset with abundant nouns. For the training time, 1$\times$ schedule denotes 12 epochs and we report the approximate number of schedules in integer format. AP$_{r}$ is the main metric and is highlighted.}
  \centering
  \resizebox{\linewidth}{!}{
      \begin{tabular}{l|c|c|c|c|cccc|cccc}
        \hline
        \multirow{2}{*}{Method} & \multirow{2}{*}{DETR} & Training & External & \multirow{2}{*}{Strict} & \multicolumn{4}{c|}{OV-LVIS Detection}  & \multicolumn{4}{c}{OV-LVIS Segmentation} \\
        & & Time & Dataset & & AP$^{\text{box}}$ & AP$^{\text{box}}_{\text{f}}$ & AP$^{\text{box}}_{\text{c}}$ & AP$^{\text{box}}_{\text{r}}$  & AP$^{\text{mask}}$ & AP$^{\text{mask}}_{\text{f}}$ & AP$^{\text{mask}}_{\text{c}}$ & AP$^{\text{mask}}_{\text{r}}$ \\
        \hline
        ViLD-Ens.~\cite{vild} & & 2$\times$ &  & \checkmark & \textcolor{gray}{27.8} & \textcolor{gray}{34.2} & \textcolor{gray}{26.5} & 16.7 & \textcolor{gray}{25.5} & \textcolor{gray}{30.3} & \textcolor{gray}{24.6} & 16.6 \\
        DetPro~\cite{detpro} & & 2$\times$ &  & \checkmark & \textcolor{gray}{28.4} & \textcolor{gray}{32.4} & \textcolor{gray}{27.8} & 20.8 & \textcolor{gray}{25.9} & \textcolor{gray}{28.9} & \textcolor{gray}{25.6} & 19.8 \\
        CondHead~\cite{wang2023learning} & & 2$\times$ &  & \checkmark & \textcolor{gray}{28.8} & \textcolor{gray}{33.7} & \textcolor{gray}{28.3} & 18.8 & \textcolor{gray}{26.4} & \textcolor{gray}{29.9} & \textcolor{gray}{26.2} & 19.1 \\
        F-VLM~\cite{fvlm} & & 8$\times$ &  & \checkmark & - & - & - & - & \textcolor{gray}{24.2} & \textcolor{gray}{26.9}& \textcolor{gray}{24.0} & 18.6 \\
        OADP~\cite{oadp} & & 2$\times$ &  & \checkmark & \textcolor{gray}{28.7} & \textcolor{gray}{32.0} & \textcolor{gray}{28.4} & 21.9 & \textcolor{gray}{26.6} & \textcolor{gray}{29.0} & \textcolor{gray}{26.3} & 21.7 \\
        BARON~\cite{BARON} & & 2$\times$ &  & \checkmark & \textcolor{gray}{28.4} & \textcolor{gray}{32.2} & \textcolor{gray}{28.4} & 20.1 & \textcolor{gray}{26.5} & \textcolor{gray}{29.4} & \textcolor{gray}{26.8} & 19.2 \\
        SAS-Det~\cite{zhao2024taming} & & 2$\times$ &  & \checkmark & - & - & - & - & \textcolor{gray}{27.4} & \textcolor{gray}{31.6} & \textcolor{gray}{26.1} & 20.9 \\
        RALF~\cite{kim2024retrieval} & & 2$\times$ &  & \checkmark & - & - & - & - & \textcolor{gray}{26.3} & \textcolor{gray}{29.2} & \textcolor{gray}{25.7} & 21.1 \\
        ProxyDet~\cite{jeong2024proxydet} & & 4$\times$ &  & \checkmark & - & - & - & - & \textcolor{gray}{30.1} & - & - & 18.9 \\
        LP-OVOD~\cite{pham2024lp} & & 2$\times$ &  & \checkmark & - & - & - & - & \textcolor{gray}{26.2} & \textcolor{gray}{29.4} & \textcolor{gray}{26.1} & 19.3 \\
        RALF~\cite{kim2024retrieval} & & 2$\times$ &  & \checkmark & - & - & - & - & \textcolor{gray}{26.3} & \textcolor{gray}{29.2} & \textcolor{gray}{25.7} & 21.1 \\
        OV-DETR~\cite{ov-detr} & \checkmark & 4$\times$ &  & \XSolidBrush & - & - & - & - & \textcolor{gray}{26.6} & \textcolor{gray}{32.5} & \textcolor{gray}{25.0} & 17.4 \\
        DK-DETR~\cite{dk-detr} & \checkmark & 6$\times$ &  & \checkmark & \textcolor{gray}{33.5} & \textcolor{gray}{40.2} & \textcolor{gray}{32.0} & 22.2 & \textcolor{gray}{30.0} & \textcolor{gray}{35.4} & \textcolor{gray}{28.9} & 20.5 \\
        \rowcolor{Gray} HD-OVD (AdaMixer, Ours) & \checkmark & 1$\times$ &  & \checkmark & \textcolor{gray}{33.1} & \textcolor{gray}{37.8} & \textcolor{gray}{32.2} & 24.7 & \textcolor{gray}{29.3} & \textcolor{gray}{33.3} & \textcolor{gray}{28.8} & 21.3 \\
        \hline
        object-centric ovd~\cite{object-centric-ovd} & & 8$\times$ & IN-L & \XSolidBrush & \textcolor{gray}{27.4} & \textcolor{gray}{31.1} & \textcolor{gray}{26.3} & 21.6 & \textcolor{gray}{25.9} & \textcolor{gray}{29.1} & \textcolor{gray}{25.0} & 21.1 \\
        PromptDet~\cite{feng2022promptdet} & & 6$\times$ & LAION-Novel & \XSolidBrush & \textcolor{gray}{27.3} & \textcolor{gray}{32.6} & \textcolor{gray}{24.7} &  21.8 & \textcolor{gray}{25.3} & \textcolor{gray}{29.3} & \textcolor{gray}{23.3} & 21.4 \\
        VLDet~\cite{vldet} & & 8$\times$ & CC3M & \checkmark & \textcolor{gray}{33.4} & \textcolor{gray}{38.7} & \textcolor{gray}{32.8} & 22.9 & \textcolor{gray}{30.1} & \textcolor{gray}{34.3} & \textcolor{gray}{29.8} & 21.7 \\
        DVDet~\cite{jin2024llms} & & 8$\times$ & CC3M & \checkmark & - & - & - & - & \textcolor{gray}{31.2} & \textcolor{gray}{35.4} & \textcolor{gray}{31.2} & 23.1 \\
        GOAT~\cite{goat} & & 8$\times$ & CC3M & \checkmark & - & - & - & - & \textcolor{gray}{30.4} & \textcolor{gray}{34.3} & \textcolor{gray}{29.7} & 23.3 \\
        CoDet~\cite{ma2023codet} & & 8$\times$ & CC3M & \checkmark & \textcolor{gray}{35.0} & \textcolor{gray}{40.1} & \textcolor{gray}{33.9} & 26.1 & \textcolor{gray}{30.7} & \textcolor{gray}{34.6} & \textcolor{gray}{30.0} & 23.4 \\
        Detic (CenterNet2)~\cite{detic} & & 8$\times$ & IN-L & \XSolidBrush & \textcolor{gray}{36.2} & \textcolor{gray}{40.3} & \textcolor{gray}{36.4} & 26.7 & \textcolor{gray}{32.4} & \textcolor{gray}{35.6} & \textcolor{gray}{32.5} & 24.6 \\
        Detic (D-DETR)~\cite{detic} & \checkmark & 8$\times$ & IN-L & \XSolidBrush & \textcolor{gray}{32.5} & \textcolor{gray}{36.6} & \textcolor{gray}{31.3} & 26.2 & - & - & - & - \\
        \rowcolor{Gray} HD-OVD (AdaMixer, Ours) & \checkmark & 2$\times$ & IN-L & \XSolidBrush & \textcolor{gray}{32.5} & \textcolor{gray}{36.3} & \textcolor{gray}{31.1} & \textbf{27.7} & \textcolor{gray}{29.1} & \textcolor{gray}{32.3} & \textcolor{gray}{27.8} & \textbf{24.9} \\
        \hline
      \end{tabular}
    }
  \label{tab:main_result_mask}
\end{table*}

\subsection{Training and Inference}
\label{train}

Combining the various distillation losses and the vanilla box regression loss, the total training loss is carried out by:
\begin{equation}
\begin{split}
    \mathcal{L} = \sum_{i=1}^{L}(\mathcal{L}_{cls}^{(i)} + \mathcal{L}_{box}^{(i)} + \alpha_1\mathcal{L}_{ckd}^{ins}\textbf{}^{(i)} + \alpha_2\mathcal{L}_{rkd}^{ins}\textbf{}^{(i)}) + \alpha_3\mathcal{L}_{ckd}^{img},
\end{split}
\end{equation}
where $\alpha_*$ is the loss coefficient, and the loss is applied to all $L$ decoder layers. Pseudo boxes do not contribute to the calculation of box losses.

During inference, the classifier is made up of both base and novel text embeddings $W_{cls}^{eval} = \{ E_{base}^T, E_{novel}^T \}$. We ensemble the scores from visual space and text space to enhance the performance following previous works~\cite{vild, fang2023simple} since the information is complementary to each other:
\begin{equation}
  p_{i,c}^{ens} = \begin{cases} 
        (\hat{p}_{i,c}^{cls})^{(1-\beta_1)}(\hat{p}_{i,c}^{dis})^{\beta_1}, \quad & \text{if }c \in C_{base}, \\
        (\hat{p}_{i,c}^{cls})^{(1-\beta_2)}(\hat{p}_{i,c}^{dis})^{\beta_2}, \quad & \text{if }c \in C_{novel}, \\
  \end{cases}
\end{equation}
where $\beta_1$ and $\beta_2$ are tunable hyperparameters. $\hat{p}_{i,c}^{cls} = \text{sigmoid}(\cos{(\hat{q}_i, e^t_j)} \cdot \tau_{cls})$ and $\hat{p}_{i,c}^{dis} = \text{sigmoid}(\cos{(q_i, e^t_j)} \cdot \tau_{ckd})$ use different object queries $\hat{q}_i$ and $q_i$. $e^t_j$ is the class embedding from $W_{cls}^{eval}$.

\section{Experiments}
\label{sec:experiment}

\subsection{Datasets}

We follow previous works to conduct experiments on \textbf{OV-LVIS}~\cite{gupta2019lvis} and \textbf{OV-COCO}~\cite{coco} datasets and test the transfer ability of the model trained on OV-LVIS to COCO~\cite{coco} and Objects365~\cite{shao2019objects365} dataset. We follow a \textbf{strict} setting where novel class names are unknown during the training unless otherwise specified. All results are presented in percentage form (\%).

\noindent\textbf{- OV-COCO}. We follow previous works to split the COCO dataset into 48 base categories and 17 novel categories, removing the remaining 15 categories. The key evaluation metric is the box AP50 of novel categories AP$_{n}$. 

\noindent\textbf{- OV-LVIS}. The OV-LVIS dataset is derived from the LVIS v1 dataset, which has 1203 categories with a long-tail distribution. Based on the category frequency, the total categories are divided into rare, common, and frequent classes. In the open-vocabulary setting, 337 rare classes are viewed as novel classes, and the remaining 866 classes, the combination of common and frequent classes, are used as base classes. During the training, novel class annotations are removed while all classes are used for testing. We report the box and mask precision AP, AP$_{r}$, AP$_{c}$ and AP$_{f}$. The AP$_{r}$ is the main metric.

\noindent\textbf{- Transfer Datasets}. Following previous works, we choose COCO and Objects365 datasets to evaluate cross-dataset generalization ability. Objects365 is a large-scale object detection dataset with 365 object categories. As for the COCO dataset, all 80 categories are used. Only the validation set is used for the generalization evaluation.

\begin{table}[t]
  \caption{Generalization ability on other datasets. We evaluate the model based on AdaMixer and trained with OV-LVIS on the validation set of COCO and Objects365 for comparisons of generalization performance.}
  \centering
  \begin{tabular}{l|ccc|ccc}
    \hline
    \multirow{2}{*}{Methods}  & \multicolumn{3}{c|}{Objects365 v2}  & \multicolumn{3}{c}{COCO} \\
     & AP & AP$_{50}$ & AP$_{75}$ & AP & AP$_{50}$ & AP$_{75}$ \\
    \hline
    ViLD~\cite{vild} & 11.8 & 18.2 & 12.6 & 36.6 & 55.6 & 39.8 \\
    F-VLM~\cite{fvlm} & 11.9 & 19.2 & 12.6 & 32.5 & 53.1 & 34.6  \\
    DetPro~\cite{detpro} & 12.1 & 18.8 & 12.9 & 34.9 & 53.8 & 37.4 \\
    DK-DETR~\cite{dk-detr} & 12.4 & 17.3 & 13.4 & 39.4 & 54.3 & 43.0 \\
    LP-OVOD~\cite{pham2024lp} & 12.6 & 18.9 & 13.1 & - & - & -  \\
    EdaDet~\cite{shi2023edadet} & 13.6 & 19.8 & 14.6 & - & - & -  \\
    BARON~\cite{BARON} & 13.6 & 21.0 & 14.5 & 36.2 & 55.7 & 39.1 \\
    \rowcolor{Gray} HD-OVD (Ours) & 13.6 & 19.9 & 14.6 & 36.6 & 53.3 & 39.6 \\
    \hline
  \end{tabular}
  \label{tab:transfer_result}
\end{table}

\subsection{Implementation Details}

HD-OVD is initially built on a DETR-like detector AdaMixer~\cite{adamixer} with 300 queries due to its fast convergence and small computation and memory cost. To demonstrate that the Hierarchical Semantic Distillation Framework is transferable to other types of detectors, we also implement HD-OVD with the well-known Faster-RCNN~\cite{ren2015faster}. And we only replace the object queries in Figure\ref{fig:model_overview} as RoI features.
We use ImageNet pre-trained ResNet-50~\cite{resnet} or Swin-B~\cite{liu2021swin} as the backbone and the CLIP ViT-B/32~\cite{clip} as the distillation teacher.

For pseudo boxes, we use the same class-agnostic proposals as previous methods~\cite{object-centric-ovd, BARON}. They are the top 5 proposals per image generated by MAVL~\cite{mavl}. Considering that some proposals may overlap with base class ground truth boxes, we simply discard them. No novel class information or annotations are leaked during this procedure. For the caption model, we use ClipCap~\cite{mokady2021clipcap}, which takes CLIP (ViT-B/32) region features as prefixes and generates captions word by word. The model is trained on the widely used image-text pair dataset CC3M~\cite{cc3m}. No novel class information leaks, and no additional bounding box annotations are used. Since all these models are frozen, we pre-extract all CLIP embeddings before training to save training computation cost.

For the model with AdaMixer, we train it using the AdamW optimizer with batch size 16 and learning rate $2.5e^{-5}$, which is the same as the base model without further tuning. For HD-OVD with Faster-RCNN, we train it using SGD optimizer with batch size 16 and learning rate 0.02, also the same as the base model. The loss coefficients $\alpha_{1}$, $\alpha_{2}$, and $\alpha_{3}$ are set to 0.5, 5.0, and 0.2. $\tau_{ckd}$, $\tau_{rkd}$ and $\tau_{cls}$ are set to 20, 5 and 50. For the OV-COCO dataset, the models are trained with 12 epochs (1$\times$ schedule) and a single-scale image size. $\beta_1$ and $\beta_2$ are set to 0.35 and 0.65. For the OV-LVIS dataset, the models are also trained with 12 epochs (1$\times$ schedule). The training image sizes range from 480 to 800. $\beta_1$ and $\beta_2$ are set to 0.25 and 0.45. Following previous works, we also use the IN-L~\cite{deng2009imagenet} dataset, which is a subset from ImageNet containing 997 classes overlapping with LVIS, to boost the performance on the OV-LVIS benchmark. We finetune the trained model mentioned above for another 1$\times$ schedule.

\subsection{Main Results}

\begin{table*}[t]
  \caption{Ablation studies on each component on OV-COCO and OV-LVIS.}
  \centering
  \begin{tabular}{cc|cc|c|c|ccc|cccc}
    \hline
    \multicolumn{2}{c|}{instance} & \multicolumn{2}{c|}{class} & image & Ens. & \multicolumn{3}{c|}{OV-COCO} &  \multicolumn{4}{c}{OV-LVIS Detection} \\
    CKD & RKD & Cls. & Wei. & ~ & & AP & AP$_b$ & AP$_n$ & AP$^{\text{box}}$ & AP$^{\text{box}}_{\text{f}}$ & AP$^{\text{box}}_{\text{c}}$ & AP$^{\text{box}}_{\text{r}}$ \\
    \hline
    & & & & & & \textcolor{gray}{46.3} & \textcolor{gray}{61.8} & 2.8 & \textcolor{gray}{33.0} & \textcolor{gray}{39.8} & \textcolor{gray}{33.2} & 17.2 \\
    \checkmark & & & & & & \textcolor{gray}{47.8} & \textcolor{gray}{61.1} & 10.3 & \textcolor{gray}{33.3} & \textcolor{gray}{39.4} & \textcolor{gray}{33.2} & 20.1 \\
    \checkmark & \checkmark & & & & & \textcolor{gray}{48.7} & \textcolor{gray}{61.6} & 12.4 & \textcolor{gray}{33.4} & \textcolor{gray}{39.0} & \textcolor{gray}{33.3} & 20.8 \\
    \hline
    \checkmark & \checkmark & \checkmark & & & & \textcolor{gray}{55.4} & \textcolor{gray}{60.2} & 41.8 & \textcolor{gray}{32.2} & \textcolor{gray}{37.9} & \textcolor{gray}{31.8} & 20.3 \\
    \checkmark & \checkmark & \checkmark & \checkmark & & & \textcolor{gray}{56.0} & \textcolor{gray}{60.7} & 43.0 & \textcolor{gray}{32.7} & \textcolor{gray}{38.2} & \textcolor{gray}{32.2} & 21.4 \\
    \hline
    \checkmark & \checkmark & \checkmark & \checkmark & \checkmark & & \textcolor{gray}{56.5} & \textcolor{gray}{61.1} & 43.5 & \textcolor{gray}{32.9} & \textcolor{gray}{38.0} & \textcolor{gray}{32.3} & 22.8 \\
    \hline
    \rowcolor{Gray} \checkmark & \checkmark & \checkmark & \checkmark & \checkmark & \checkmark & \textcolor{gray}{57.4} & \textcolor{gray}{61.3} & 46.4 & \textcolor{gray}{33.1} & \textcolor{gray}{37.8} & \textcolor{gray}{32.2} & 24.7 \\
    \hline
  \end{tabular}
  \label{tab:ab1}
\end{table*}

\begin{table*}[t]
\vspace{-0.8em}
\centering
\begin{minipage}{0.3\linewidth}{
    \begin{center}
    \caption{Ablation studies on distillation classes on OV-COCO.}
        \begin{tabular}{cc|ccc}
            \hline
            base & pseudo & AP & AP$_b$ & AP$_n$ \\
            \hline
            \checkmark &  & \textcolor{gray}{46.9} & \textcolor{gray}{61.6} & 5.5 \\
            & \checkmark & \textcolor{gray}{55.8} & \textcolor{gray}{60.6} & 42.4 \\
            \rowcolor{Gray} \checkmark & \checkmark & \textcolor{gray}{57.3} & \textcolor{gray}{61.4} & 45.9 \\
            \hline
          \end{tabular}\label{tab:ab3}
    \end{center}}
\end{minipage}
\hspace{0.5em}
\begin{minipage}{0.3\linewidth}{
    \begin{center}
    \caption{Ablation studies on caption models on OV-COCO.}
        \begin{tabular}{l|ccc}
            \hline
            Caption Model & AP & AP$_b$ & AP$_n$ \\
            \hline
            \rowcolor{Gray} ClipCap~\cite{mokady2021clipcap} & \textcolor{gray}{57.3} & \textcolor{gray}{61.4} & 45.9 \\
            BLIP-2~\cite{li2023blip} & \textcolor{gray}{57.8} & \textcolor{gray}{61.4} & 47.7 \\
            \hline
          \end{tabular}\label{tab:sab4}
    \end{center}}
\end{minipage}
\hspace{0.5em}
\begin{minipage}{0.34\linewidth}{
    \begin{center}
    \caption{Ablation studies on pseudo text label generation pipeline on OV-COCO.}
        \begin{tabular}{l|ccc}
            \hline
            Pseudo Text Label & AP & AP$_b$ & AP$_n$ \\
            \hline
            \rowcolor{Gray} Processed Nouns & \textcolor{gray}{57.3} & \textcolor{gray}{61.4} & 45.9 \\
            Raw Texts & \textcolor{gray}{56.2} & \textcolor{gray}{61.0} & 42.7 \\
            \hline
          \end{tabular} \label{tab:sab5}
    \end{center}}
\end{minipage}
\vspace{-0.8em}
\end{table*}

\noindent\textbf{- OV-COCO benchmark.} As shown in Table~\ref{tab:main_result_coco}, our method with AdaMixer achieves an impressive 46.4\% AP$_{n}$ without leaking novel class information or using external datasets, outperforming previous methods by 5\% to 10\% AP. Note that our method will not hurt the base class performance and also get a high AP$_{b}$. Moreover, compared with F-VLM~\cite{fvlm} and CORA~\cite{wu2023cora} which use frozen CLIP image encoder as the backbone to preserve its generalization ability, our method can switch to different backbones, not limited to the pre-trained VLMs. For example, HD-OVD with Swin-B can further scale to 52.7\% AP$_{n}$. Our HD-OVD also enjoys high efficiency during inference without needing additional forwards in the CLIP image encoder like EdaDet~\cite{shi2023edadet} or the CLIP text encoder like BARON~\cite{BARON}. In addition, HD-OVD enjoys high adaptability. The model with Faster-RCNN as the base model also achieves 45.2\% AP$_{n}$ and 46.3\% AP$_{n}$ under $1\times$ and $2\times$ training time, respectively.

\noindent\textbf{- OV-LVIS benchmark.} In the left part of Table~\ref{tab:main_result_mask}, our HD-OVD also achieves SOTA performance on the OV-LVIS benchmark with a 24.7\% AP$_r^{box}$ without using additional datasets or annotations. The LVIS dataset is a federated dataset in which not all classes are annotated for each image. We find that a large proportion of the pseudo boxes are actually miss-annotated base classes. Using IN-L dataset to explore more diverse images and objects can further boost the AP$_r^{box}$ to 27.7\% AP, outperforming other methods using additional datasets by a clear margin.

The right part of Table~\ref{tab:main_result_mask} shows the mask results on the OV-LVIS dataset compared with other methods. Since there is no mask head tailored for AdaMixer, we add an additional convolutional mask head to the detector and use a two-stage training recipe following DETR~\cite{detr}, in which we fix the detector and only train the mask head for 1$\times$ schedule. As shown in Table~\ref{tab:main_result_mask}, our method can easily extend to instance segmentation. The generalization ability for novel classes is maintained and we achieve the highest 24.9\% AP$^{\text{mask}}_{\text{r}}$.

\noindent\textbf{- Generalization ability on other datasets.} To simulate real-world applications, we conduct cross-dataset generalization ability testing on the Objects356 dataset and COCO dataset with the model training on the OV-LVIS dataset without using IN-L. Following previous practices, we only replace the text embeddings in the classifier with the target dataset categories without additional finetuning. As shown in Table~\ref{tab:transfer_result}, our method also achieves impressive 13.6\% AP and 36.6\% AP on Objects365 and COCO. Our method can achieve high performance on both datasets, showing its great generalization ability on different domains and vocabularies.

\subsection{Ablation Studies}

In this subsection, we use AdaMixer as the base model. Models are trained with 12 epochs without the IN-L dataset or mask annotations. Rows in \colorbox{Gray}{gray} are default settings.

\noindent\textbf{- Effect of image-space instance-wise distillation.} As shown in the first block in Table~\ref{tab:ab1}, the model only trained with base annotations has limited generalization ability with 2.8\% and 17.2\% novel AP on OV-COCO and OV-LVIS datasets, respectively. Using contrastive knowledge distillation in visual space to align object features with fine-grained CLIP region features boosts the AP for novel classes to 10.3\% and 20.1\%. Further, cooperating with relational distillation to model relations among objects yields a 2.1\% and 0.7\% increase in novel classes AP.

\begin{figure*}[tb]
  \centering
  \includegraphics[width=0.9\linewidth]{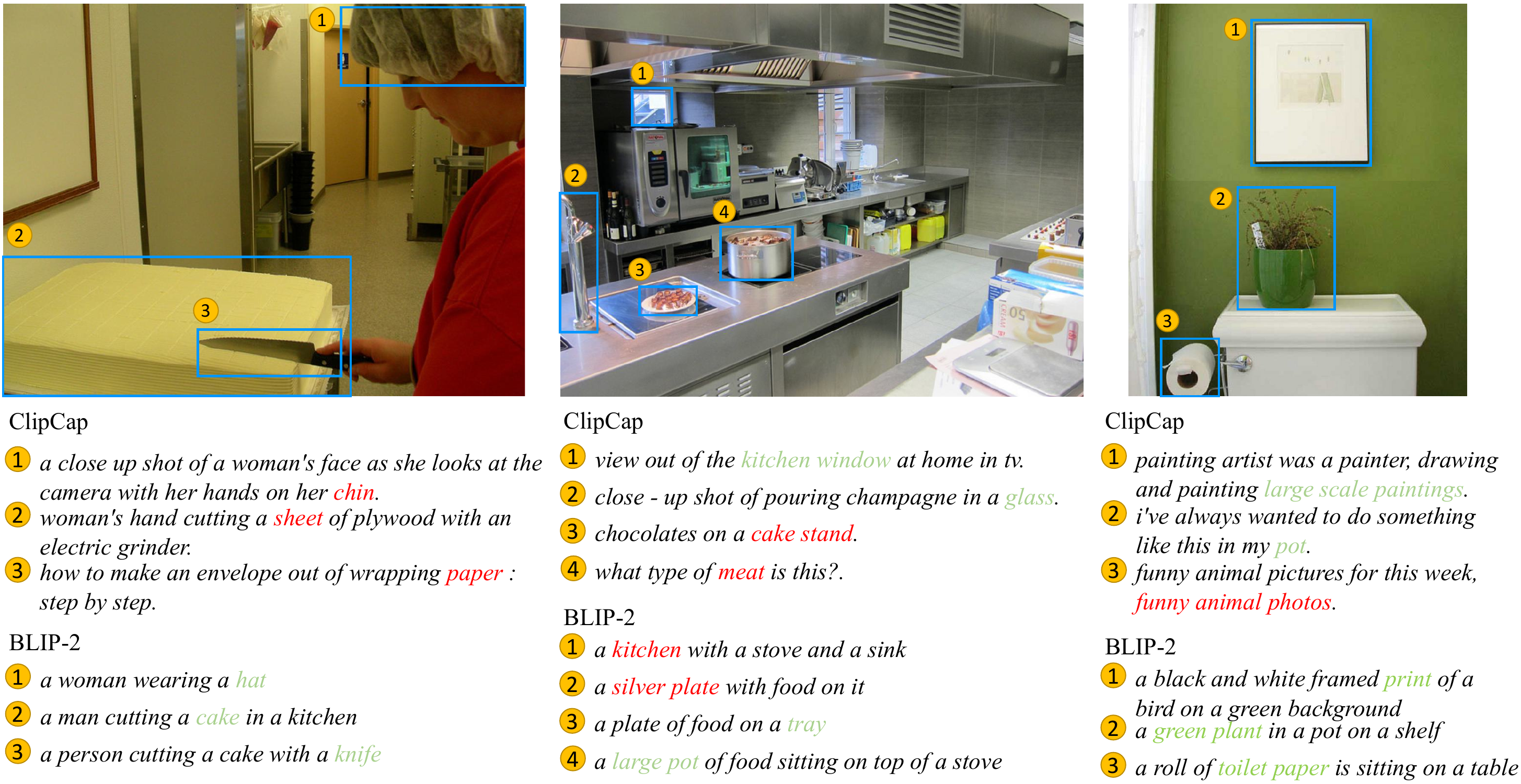}
  \vspace{-0.6em}
  \caption{Visualization of generated captions and pseudo labels. Pseudo boxes are marked in \textcolor{blue}{blue} boxes. The correct and incorrect pseudo labels are colored in \textcolor{green}{green} and \textcolor{red}{red}. The powerful BLIP-2 model can generate more accurate pseudo text labels than ClipCap.}
  \vspace{-0.6em}
  \label{fig:captions}
\end{figure*}

\noindent\textbf{- Effect of text-space class-wise distillation.} Based on instance-wise distillation in image space, our HD-OVD further generates pseudo text labels for each pseudo box.
This allows us to associate pseudo boxes with corresponding text labels instead of categorizing them as background during the classification.
As shown in the second block in Table~\ref{tab:ab1}, if we do not decrease the negative impacts from noisy labels, not only novel AP but also base AP will be affected. Figure~\ref{fig:captions} visualizes some generated pseudo text labels, showing that the pseudo text labels are not always acceptable. But with proper weighting, the novel AP can achieve 43.0\% AP and 21.4\% AP. Although the LVIS dataset is a federated dataset and a large proportion of the pseudo boxes are actually miss-annotated base classes, assigning them with synonyms text labels is still helpful. Since the OV-COCO dataset is fully annotated, the gains from pseudo text labeling are more significant.

\noindent\textbf{- Effect of image-wise contrastive distillation and score ensemble.} Since the whole image contains multiple objects and rich contexts, learning global image features from CLIP explores a comprehensive understanding for different images, which increases 0.5\% AP$_n$ and 1.4\% AP$_r$. Further, since instance-wise knowledge from the CLIP image encoder and class-wise knowledge from the CLIP text encoder are complementary, ensembling the scores can boost the novel AP to the final 46.4\% and 24.7\%.

\noindent\textbf{- Effect of different distillation classes.} In this work, we explicitly model the novel-base and novel-background relationships using some pseudo boxes. As shown in Table~\ref{tab:ab3}, only distilling on base classes can not improve the generalization ability since base classes are supervised under gold human annotations. Distilling with only pseudo novel classes can get complementary knowledge aside from base class annotations thus achieving a remarkable novel AP. Further, considering the relationships among novel classes, base classes, and background carefully can boost the novel AP by 3.5\%, making them easily distinguishable.

\noindent{\textbf{- Effect of a stronger caption model.}} In HD-OVD, we use an off-the-shelf caption model to generate descriptions for each class-agnostic pseudo box, from which we extract the noun phrases as pseudo text labels. A stronger caption model can generate more accurate pseudo-text labels. By default, we use ClipCap~\cite{mokady2021clipcap} as the caption model, which uses CLIP-ViT-b/32~\cite{clip} as the image encoder and GPT-2~\cite{gpt2} as the language model. In this ablation study, we use a state-of-the-art caption model BLIP-2~\cite{li2023blip}. BLIP-2, containing around 4.1B parameters, uses ViT-g/14 as the image encoder and FlanT5-xl~\cite{Flant5} as the language model. As shown in Table~\ref{tab:sab4}, using BLIP-2 can increase 1.8\% AP$_n$. Figure~\ref{fig:captions} visualizes the generated captions and the extracted pseudo text labels from ClipCap and BLIP-2, separately, showing that the pseudo text labels generated by BLIP-2 are more accurate and a stronger caption model can further boost the detection performance.

\noindent{\textbf{- Effect of pseudo text label generation pipeline.}} In HD-OVD, we design an automatic pipeline to extract the most suitable nouns in captions as pseudo-text labels. If we directly use the generated raw texts as the pseudo text labels, the novel AP will decrease by 3.2\% AP, as shown in Table~\ref{tab:sab5}.

\noindent{\textbf{- Effect of maximum number of pseudo boxes.}} By default, HD-OVD uses the top-5 proposals as pseudo boxes. Proposals overlapping with base ground truth boxes are discarded. In Table~\ref{tab:r5}, we decrease the number of pseudo labels linearly and the performance also decreases accordingly. When we do not use pseudo boxes totally, the distillation is only performed on base classes (the results in Table~\ref{tab:ab3}).

\begin{table}[t]
\vspace{-0.4em}
  \caption{Ablation studies on maximum number of pseudo boxes on OV-COCO.}
  \vspace{-0.4em}
  \centering
  \begin{tabular}{c|ccc}
    \hline
    Number & AP & AP$_b$ & AP$_n$ \\
    \hline
    0 & \textcolor{gray}{46.9} & \textcolor{gray}{61.6} & 5.5 \\
    1 & \textcolor{gray}{57.2} & \textcolor{gray}{61.6} & 45.0 \\
    2 & \textcolor{gray}{57.5} & \textcolor{gray}{61.6} & 45.9 \\
    3 & \textcolor{gray}{57.8} & \textcolor{gray}{62.0} & 46.1 \\
    4 & \textcolor{gray}{57.5} & \textcolor{gray}{61.5} & 46.3 \\
    \rowcolor{Gray} 5 & \textcolor{gray}{57.4} & \textcolor{gray}{61.3} & \textbf{46.4} \\
    \hline
  \end{tabular}
  \label{tab:r5}
  \vspace{-0.4em}
\end{table}

\noindent{\textbf{- Effect of different distillation pipelines.}} Our HD-OVD performs instance-wise distillation and class-wise distillation in a separate and progressive manner, by first aligning with the CLIP image encoder and then learning from the CLIP text encoder. Table~\ref{tab:sab3} validates the superiority of this design. If we distill the dual encoders in parallel, the novel AP will slightly degrade by 0.6\% AP. However, if the dual distillation is performed in the same space without using a projector to separate the space, the novel AP will decrease by 2.1\% AP, demonstrating that the supervision signals from different encoders may be conflicting.

\begin{table}[t]
  \caption{Ablation studies on distillation pipeline on OV-COCO.}
  \vspace{-0.4em}
  \centering
  \begin{tabular}{l|ccc}
    \hline
    Method & AP & AP$_b$ & AP$_n$ \\
    \hline
    \rowcolor{Gray} Progressive and Separate & \textcolor{gray}{57.3} & \textcolor{gray}{61.4} & 45.9 \\
    Parallel and Separate & \textcolor{gray}{57.4} & \textcolor{gray}{61.7} & 45.3 \\
    In the same space & \textcolor{gray}{56.3} & \textcolor{gray}{60.8} & 43.8 \\
    \hline
  \end{tabular}
  \vspace{-0.4em}
  \label{tab:sab3}
\end{table}

\noindent{\textbf{- Effect of training hyperparameters.}} Other hyperparameters include the temperature coefficients $\tau_{cls}$, $\tau_{ckd}$, $\tau_{rkd}$, loss coefficients $\alpha_1$, $\alpha_2$, $\alpha_3$, and the number of padded negative samples in contrastive knowledge distillation queue\_len (ins) and queue\_len (img). As shown in Table~\ref{tab:R1}, our default hyperparameters achieve good novel-base performance trade-off and the model is robust to the choice of hyperparameters.

\begin{table}[t]
  \caption{Ablation studies on training hyperparameters on OV-COCO.}
  \vspace{-0.4em}
  \centering
  \resizebox{\linewidth}{!}{
  \begin{tabular}{|c|c||c|c|}
    \hline
     $\tau_{cls}$ & AP/AP$_b$/AP$_n$ & $\alpha_1$ & AP/AP$_b$/AP$_n$ \\
    \hline
    100 & \textcolor{gray}{57.3}/\textcolor{gray}{61.4}/45.8 & 1.0 & \textcolor{gray}{57.2}/\textcolor{gray}{61.1}/46.3 \\
    \rowcolor{Gray} 50 & \textcolor{gray}{57.3}/\textcolor{gray}{61.4}/45.9 & 0.5 & \textcolor{gray}{57.3}/\textcolor{gray}{61.4}/45.9 \\
    20 & \textcolor{gray}{56.7}/\textcolor{gray}{60.8}/45.0 & 0.3 & \textcolor{gray}{57.0}/\textcolor{gray}{60.9}/46.0 \\
    \hline
    \hline
    $\tau_{ckd}$ & AP/AP$_b$/AP$_n$ & $\alpha_2$ & AP/AP$_b$/AP$_n$ \\
    \hline
    50 & \textcolor{gray}{56.9}/\textcolor{gray}{61.0}/45.4 & 10 & \textcolor{gray}{57.3}/\textcolor{gray}{61.3}/45.9 \\
    \rowcolor{Gray} 20 & \textcolor{gray}{57.3}/\textcolor{gray}{61.4}/45.9 & 5 & \textcolor{gray}{57.3}/\textcolor{gray}{61.4}/45.9 \\
    10 & \textcolor{gray}{56.8}/\textcolor{gray}{61.4}/44.0 & 3 & \textcolor{gray}{57.4}/\textcolor{gray}{61.5}/46.0 \\
    \hline
    \hline
    $\tau_{rkd}$ & AP/AP$_b$/AP$_n$ & $\alpha_3$ & AP/AP$_b$/AP$_n$ \\
    \hline
    10 & \textcolor{gray}{57.1}/\textcolor{gray}{61.2}/45.5 & 0.3 & \textcolor{gray}{57.5}/\textcolor{gray}{61.6}/46.0 \\
    \rowcolor{Gray} 5 & \textcolor{gray}{57.3}/\textcolor{gray}{61.4}/45.9 & 0.2 & \textcolor{gray}{57.3}/\textcolor{gray}{61.4}/45.9 \\
    3 & \textcolor{gray}{57.3}/\textcolor{gray}{61.5}/45.5 & 0.1 & \textcolor{gray}{57.1}/\textcolor{gray}{61.1}/45.9 \\
    \hline
    \hline
    queue\_len (ins) & AP/AP$_b$/AP$_n$ & queue\_len (img) & AP/AP$_b$/AP$_n$ \\
    \hline
    \cellcolor{Gray} 2048 & \cellcolor{Gray} \textcolor{gray}{57.5}/\textcolor{gray}{61.6}/46.0 & \cellcolor{white} 2048 & \cellcolor{white} \textcolor{gray}{57.2}/\textcolor{gray}{61.2}/46.1 \\
    \cellcolor{white} 1024 & \cellcolor{white} \textcolor{gray}{57.3}/\textcolor{gray}{61.4}/45.9 & \cellcolor{white} 1024 & \cellcolor{white} \textcolor{gray}{57.3}/\textcolor{gray}{61.4}/45.9 \\
    \cellcolor{white} 512 & \cellcolor{white} \textcolor{gray}{57.1}/\textcolor{gray}{61.1}/46.0 & \cellcolor{Gray} 512 & \cellcolor{Gray} \textcolor{gray}{57.5}/\textcolor{gray}{61.6}/45.9 \\
    \hline
  \end{tabular}}
  \vspace{-0.4em}
  \label{tab:R1}
\end{table}

\noindent{\textbf{- Effect of different ensemble coefficients.}} We ensemble the classification scores from visual space and text space to boost the performance. As shown in Table~\ref{tab:sab1}, only using text space scores (the first row) or only using visual space scores (the last row) achieves 22.8\% and 19.4\% novel AP. However, simply ensembling the complementary information from both spaces can increase the novel AP to 24.7\%. We also find that the results are robust to the ensemble coefficients.

\begin{table}[t]
  \caption{Ablation studies on ensemble coefficients on OV-LVIS.}
  \vspace{-0.4em}
  \centering
  \begin{tabular}{l|c|c|cccc}
    \hline
    Method & $\beta_1$ & $\beta_2$ & AP$^{\text{box}}$ & AP$^{\text{box}}_{\text{f}}$ & AP$^{\text{box}}_{\text{c}}$ & AP$^{\text{box}}_{\text{r}}$ \\
    \hline
    Text-only & 1.0 & 1.0 & \textcolor{gray}{32.9} & \textcolor{gray}{38.0} & \textcolor{gray}{32.3} & 22.8 \\
    \hline
    \multirow{8}{*}{Ensemble} & \multirow{5}{*}{0.35} & 0.75 & \textcolor{gray}{31.4} & \textcolor{gray}{35.6} & \textcolor{gray}{30.5} & 24.3 \\
     & & 0.65 & \textcolor{gray}{32.4} & \textcolor{gray}{37.0} & \textcolor{gray}{31.6} & 24.3 \\
     & & 0.55 & \textcolor{gray}{32.9} & \textcolor{gray}{37.6} & \textcolor{gray}{32.1} & 24.4 \\
     & & 0.45 & \textcolor{gray}{33.2} & \textcolor{gray}{37.9} & \textcolor{gray}{32.4} & 24.5 \\
     & & 0.35 & \textcolor{gray}{33.3} & \textcolor{gray}{38.2} & \textcolor{gray}{32.6} & 24.1 \\
    \cline{2-7}
     & 0.35 & & \textcolor{gray}{33.2} & \textcolor{gray}{37.9} & \textcolor{gray}{32.4} & 24.5 \\
     & \cellcolor{Gray}0.25 & \cellcolor{Gray}0.45 & \cellcolor{Gray}\textcolor{gray}{33.1} & \cellcolor{Gray}\textcolor{gray}{37.8} & \cellcolor{Gray}\textcolor{gray}{32.2} & \cellcolor{Gray}24.7 \\
    & 0.15 &  & \textcolor{gray}{32.8} & \textcolor{gray}{37.5} & \textcolor{gray}{31.8} & 24.8 \\
    \hline
    Visual-only & 0.0 & 0.0 & \textcolor{gray}{25.1} & \textcolor{gray}{27.9} & \textcolor{gray}{24.7} & 19.4 \\
    \hline
  \end{tabular}
  \vspace{-0.4em}
  \label{tab:sab1}
\end{table}

\begin{figure*}[tb]
  \centering
  \vspace{-0.6em}
  \includegraphics[width=1\linewidth]{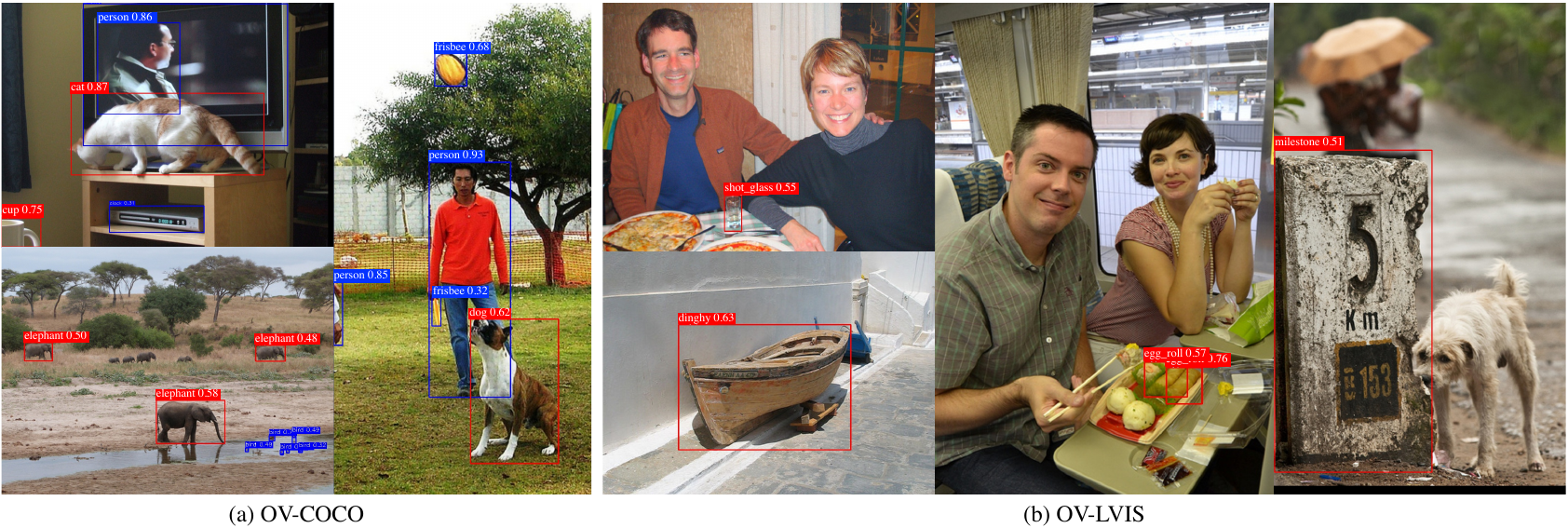}\vspace{-1em}
  \caption{Visualization of detected novel-class objects. Novel and base classes are marked in \textcolor{red}{red} and \textcolor{blue}{blue} boxes, separately. For clarity, base classes on the OV-LVIS dataset are not visualized in the pictures. Zoom in for the best view.}\vspace{-1em}
  \label{fig:vis}
\end{figure*}

\noindent\textbf{- Visualization.} Figure~\ref{fig:vis} visualizes inference results from the models trained on OV-COCO and OV-LVIS, respectively. Our model can detect novel-class objects with high confidence and accurate localization.

\section{Discussion}
\subsection{Comparisons between the Performance on OV-LVIS and OV-COCO Datasets}
Although HD-OVD achieves state-of-the-art performance on both the OV-LVIS and OV-COCO datasets, the performance gain on the OV-LVIS dataset is relatively lower than the one on the OV-COCO dataset. We find that two factors may affect the performance on OV-LVIS: \textbf{First}, we utilize top-scored proposals as pseudo boxes. However, LVIS is a federated dataset in which each image is only annotated on a subset of classes and a large proportion of the pseudo boxes are actually miss-annotated base classes. For example, the clock, oranges, and cat in Figure~\ref{fig:model_overview} are all base classes in LVIS. Thus, distilling on base classes gets relatively low improvements, validated in Table~\ref{tab:ab3}. In this work, we use an external dataset IN-L following previous works to ensure that pseudo boxes cover a broad range of concepts and improve the APr by 3\%. In contrast, the COCO dataset is fully annotated. \textbf{Second}, the categories in LVIS are much more fine-grained than the ones in COCO. For example, ``clock'' in COCO is divided into ``clock'', ``clock tower'' and ``wall clock'' in LVIS. Without carefully tuned, the caption model tends to provide coarse category labels and thus may not be sufficient for LVIS. In Table~\ref{tab:sab4}, we show that a stronger caption model can boost the performance. And recent large multi-modal models with strong perception ability can generate fine-grained labels for each pseudo box with carefully designed prompts, which is a promising alternative solution.

\begin{table}[t]
  \vspace{-0.6em}
  \caption{Comparison on different sizes of objects on OV-COCO. The AP in this table is the average AP from 0.5 to 0.95 IoU thresholds.}
  \vspace{-0.6em}
  \centering
  \resizebox{\linewidth}{!}{
  \begin{tabular}{l|c|ccc|ccc}
    \hline
    \multirow{2}{*}{Method} & Training & \multicolumn{3}{c|}{Base Class}  & \multicolumn{3}{c}{Novel Class} \\
     & Time & AP$_{s}$ & AP$_{m}$ & AP$_{l}$ & AP$_{s}$ & AP$_{m}$ & AP$_{l}$ \\
    \hline
    OADP~\cite{oadp} & 2$\times$ & 22.5 & 37.8 & 47.2 & 12.6 & 19.8 & 20.3 \\
    BARON~\cite{BARON} & 2$\times$ & 23.5 & 41.3 & 50.4 & 16.3 & 24.3 & 25.5 \\
    HD-OVD (Faster-RCNN, Ours) & 1$\times$ & 20.7 & 37.0 & 44.8 & 14.8 & 33.2 & 35.1 \\
    HD-OVD (Faster-RCNN, Ours) & 2$\times$ & 21.6 & 38.4 & 46.4 & 15.5 & 34.4 & 35.8 \\
    HD-OVD (AdaMixer, Ours) & 1$\times$ & 21.6 & 41.8 & 55.1 & 14.9 & 36.8 & 42.8 \\
    \hline
  \end{tabular}}
  \vspace{-0.6em}
  \label{tab:r3}
\end{table}

\subsection{Effect on Small Novel Objects}
Although the instance-wise distillation in HD-OVD is designed to learn fine-grained object details and the performance of large novel objects is significantly higher than other methods, we find that the performance of small novel objects is only comparable with others, as shown in Table~\ref{tab:r3}. The pseudo boxes in HD-OVD are top-scored proposals from an offline region proposal network. Thus the
pseudo boxes tend to be large and salient objects as these objects are easy to detect, making the learning biased toward large objects. Giving some priority to small object proposals when selecting pseudo boxes may help alleviate the problem. Further, data augmentations including multi-scale training or large-scale jitter may also improve the performance of small novel objects.

\section{Conclusion}

In this work, a comprehensive Hierarchical Semantic Distillation framework is proposed to learn highly distinguishable semantic knowledge for novel classes. 
This framework integrates instance-wise fine-grained relations, class-wise generalizable text labels, and complex image-level representations into a semantic hierarchy, enhancing the complementarity of the distillation.
Superior results on challenging benchmarks validate the effectiveness of our HD-OVD. We believe more diverse and accurate pseudo boxes and a more powerful caption model could further improve the generalization ability.



\section*{Acknowledgments}
This work was supported partially by the National Key Research and Development Program of China (2023YFA1008503), NSFC(92470202, U21A20471), National Natural Science Foundation of China (U22A2095), Guangdong NSF Project (No. 2023B1515040025), the Project of Guangdong Provincial Key Laboratory of Information Security Technology (2023B1212060026), the Major Key Project of PCL under Grant PCL2024A06. This work was also supported by Alibaba Innovative Research Program.



 
\bibliographystyle{IEEEtran}
\bibliography{sample-base}

\begin{thebibliography}{10}
\providecommand{\url}[1]{#1}
\csname url@samestyle\endcsname
\providecommand{\newblock}{\relax}
\providecommand{\bibinfo}[2]{#2}
\providecommand{\BIBentrySTDinterwordspacing}{\spaceskip=0pt\relax}
\providecommand{\BIBentryALTinterwordstretchfactor}{4}
\providecommand{\BIBentryALTinterwordspacing}{\spaceskip=\fontdimen2\font plus
\BIBentryALTinterwordstretchfactor\fontdimen3\font minus \fontdimen4\font\relax}
\providecommand{\BIBforeignlanguage}[2]{{%
\expandafter\ifx\csname l@#1\endcsname\relax
\typeout{** WARNING: IEEEtran.bst: No hyphenation pattern has been}%
\typeout{** loaded for the language `#1'. Using the pattern for}%
\typeout{** the default language instead.}%
\else
\language=\csname l@#1\endcsname
\fi
#2}}
\providecommand{\BIBdecl}{\relax}
\BIBdecl

\bibitem{oadp}
L.~Wang, Y.~Liu, P.~Du, Z.~Ding, Y.~Liao, Q.~Qi, B.~Chen, and S.~Liu, ``Object-aware distillation pyramid for open-vocabulary object detection,'' in \emph{Proceedings of the IEEE/CVF Conference on Computer Vision and Pattern Recognition}, 2023, pp. 11\,186--11\,196.

\bibitem{ren2015faster}
S.~Ren, K.~He, R.~Girshick, and J.~Sun, ``Faster r-cnn: Towards real-time object detection with region proposal networks,'' \emph{IEEE Transactions on Pattern Analysis and Machine Intelligence}, vol.~39, no.~6, pp. 1137--1149, 2017.

\bibitem{tian2019fcos}
Z.~Tian, C.~Shen, H.~Chen, and T.~He, ``Fcos: A simple and strong anchor-free object detector,'' \emph{IEEE Transactions on Pattern Analysis and Machine Intelligence}, vol.~44, no.~4, pp. 1922--1933, 2020.

\bibitem{detr}
N.~Carion, F.~Massa, G.~Synnaeve, N.~Usunier, A.~Kirillov, and S.~Zagoruyko, ``End-to-end object detection with transformers,'' in \emph{European Conference on Computer Vision}, 2020, pp. 213--229.

\bibitem{deformabledetr}
X.~Zhu, W.~Su, L.~Lu, B.~Li, X.~Wang, and J.~Dai, ``Deformable detr: Deformable transformers for end-to-end object detection,'' in \emph{International Conference on Learning Representations}, 2021.

\bibitem{fu2023asag}
S.~Fu, J.~Yan, Y.~Gao, X.~Xie, and W.-S. Zheng, ``Asag: Building strong one-decoder-layer sparse detectors via adaptive sparse anchor generation,'' in \emph{Proceedings of the IEEE/CVF International Conference on Computer Vision}, 2023, pp. 6328--6338.

\bibitem{9416174}
S.~Wu, Y.~Xu, B.~Zhang, J.~Yang, and D.~Zhang, ``Deformable template network (dtn) for object detection,'' \emph{IEEE Transactions on Multimedia}, vol.~24, pp. 2058--2068, 2022.

\bibitem{9684715}
J.~Xie, Y.~Pang, J.~Nie, J.~Cao, and J.~Han, ``Latent feature pyramid network for object detection,'' \emph{IEEE Transactions on Multimedia}, vol.~25, pp. 2153--2163, 2023.

\bibitem{mo2024bridge}
Q.~Mo, Y.~Gao, S.~Fu, J.~Yan, A.~Wu, and W.-S. Zheng, ``Bridge past and future: Overcoming information asymmetry in incremental object detection,'' in \emph{European Conference on Computer Vision}.\hskip 1em plus 0.5em minus 0.4em\relax Springer, 2024, pp. 463--480.

\bibitem{fu2024frozen}
S.~Fu, J.~Yan, Q.~Yang, X.~Wei, X.~Xie, and W.-S. Zheng, ``Frozen-detr: Enhancing detr with image understanding from frozen foundation models,'' \emph{arXiv preprint arXiv:2410.19635}, 2024.

\bibitem{adamixer}
Z.~Gao, L.~Wang, B.~Han, and S.~Guo, ``Adamixer: A fast-converging query-based object detector,'' in \emph{Proceedings of the IEEE/CVF Conference on Computer Vision and Pattern Recognition}, 2022, pp. 5364--5373.

\bibitem{wu2024towards}
J.~Wu, X.~Li, S.~Xu, H.~Yuan, H.~Ding, Y.~Yang, X.~Li, J.~Zhang, Y.~Tong, X.~Jiang \emph{et~al.}, ``Towards open vocabulary learning: A survey,'' \emph{IEEE Transactions on Pattern Analysis and Machine Intelligence}, 2024.

\bibitem{ovr-cnn}
A.~Zareian, K.~D. Rosa, D.~H. Hu, and S.-F. Chang, ``Open-vocabulary object detection using captions,'' in \emph{Proceedings of the IEEE/CVF Conference on Computer Vision and Pattern Recognition}, 2021, pp. 14\,393--14\,402.

\bibitem{vild}
X.~Gu, T.-Y. Lin, W.~Kuo, and Y.~Cui, ``Open-vocabulary object detection via vision and language knowledge distillation,'' in \emph{International Conference on Learning Representations}, 2022.

\bibitem{clip}
A.~Radford, J.~W. Kim, C.~Hallacy, A.~Ramesh, G.~Goh, S.~Agarwal, G.~Sastry, A.~Askell, P.~Mishkin, J.~Clark \emph{et~al.}, ``Learning transferable visual models from natural language supervision,'' in \emph{International Conference on Machine Learning}, 2021, pp. 8748--8763.

\bibitem{BARON}
S.~Wu, W.~Zhang, S.~Jin, W.~Liu, and C.~C. Loy, ``Aligning bag of regions for open-vocabulary object detection,'' in \emph{Proceedings of the IEEE/CVF Conference on Computer Vision and Pattern Recognition}, 2023, pp. 15\,254--15\,264.

\bibitem{dk-detr}
L.~Li, J.~Miao, D.~Shi, W.~Tan, Y.~Ren, Y.~Yang, and S.~Pu, ``Distilling detr with visual-linguistic knowledge for open-vocabulary object detection,'' in \emph{Proceedings of the IEEE/CVF International Conference on Computer Vision}, 2023, pp. 6501--6510.

\bibitem{vldet}
C.~Lin, P.~Sun, Y.~Jiang, P.~Luo, L.~Qu, G.~Haffari, Z.~Yuan, and J.~Cai, ``Learning object-language alignments for open-vocabulary object detection,'' in \emph{International Conference on Learning Representations}, 2023.

\bibitem{goat}
J.~Wang, H.~Zhang, H.~Hong, X.~Jin, Y.~He, H.~Xue, and Z.~Zhao, ``Open-vocabulary object detection with an open corpus,'' in \emph{Proceedings of the IEEE/CVF International Conference on Computer Vision}, 2023, pp. 6759--6769.

\bibitem{detic}
X.~Zhou, R.~Girdhar, A.~Joulin, P.~Kr{\"a}henb{\"u}hl, and I.~Misra, ``Detecting twenty-thousand classes using image-level supervision,'' in \emph{European Conference on Computer Vision}, 2022, pp. 350--368.

\bibitem{feng2022promptdet}
C.~Feng, Y.~Zhong, Z.~Jie, X.~Chu, H.~Ren, X.~Wei, W.~Xie, and L.~Ma, ``Promptdet: Towards open-vocabulary detection using uncurated images,'' in \emph{European Conference on Computer Vision}, 2022.

\bibitem{lin2017feature}
T.-Y. Lin, P.~Doll{\'a}r, R.~Girshick, K.~He, B.~Hariharan, and S.~Belongie, ``Feature pyramid networks for object detection,'' in \emph{Proceedings of the IEEE/CVF Conference on Computer Vision and Pattern Recognition}, 2017, pp. 2117--2125.

\bibitem{lin2017focal}
T.-Y. Lin, P.~Goyal, R.~Girshick, K.~He, and P.~Doll{\'a}r, ``Focal loss for dense object detection,'' in \emph{Proceedings of the IEEE/CVF International Conference on Computer Vision}, 2017, pp. 2980--2988.

\bibitem{dosovitskiy2020image}
A.~Dosovitskiy, L.~Beyer, A.~Kolesnikov, D.~Weissenborn, X.~Zhai, T.~Unterthiner, M.~Dehghani, M.~Minderer, G.~Heigold, S.~Gelly \emph{et~al.}, ``An image is worth 16x16 words: Transformers for image recognition at scale,'' \emph{arXiv preprint arXiv:2010.11929}, 2020.

\bibitem{10041780}
J.~Jiao, Y.-M. Tang, K.-Y. Lin, Y.~Gao, A.~J. Ma, Y.~Wang, and W.-S. Zheng, ``Dilateformer: Multi-scale dilated transformer for visual recognition,'' \emph{IEEE Transactions on Multimedia}, pp. 8906--8919, 2023.

\bibitem{10380775}
Y.~Lu, S.~Sirejiding, Y.~Ding, C.~Wang, and H.~Lu, ``Prompt guided transformer for multi-task dense prediction,'' \emph{IEEE Transactions on Multimedia}, vol.~26, pp. 6375--6385, 2024.

\bibitem{ma2022open}
Z.~Ma, G.~Luo, J.~Gao, L.~Li, Y.~Chen, S.~Wang, C.~Zhang, and W.~Hu, ``Open-vocabulary one-stage detection with hierarchical visual-language knowledge distillation,'' in \emph{Proceedings of the IEEE/CVF Conference on Computer Vision and Pattern Recognition}, 2022, pp. 14\,074--14\,083.

\bibitem{shi2023edadet}
C.~Shi and S.~Yang, ``Edadet: Open-vocabulary object detection using early dense alignment,'' in \emph{Proceedings of the IEEE/CVF International Conference on Computer Vision}, 2023.

\bibitem{gao2022open}
M.~Gao, C.~Xing, J.~C. Niebles, J.~Li, R.~Xu, W.~Liu, and C.~Xiong, ``Open vocabulary object detection with pseudo bounding-box labels,'' in \emph{European Conference on Computer Vision}.\hskip 1em plus 0.5em minus 0.4em\relax Springer, 2022, pp. 266--282.

\bibitem{ma2023codet}
C.~Ma, Y.~Jiang, X.~Wen, Z.~Yuan, and X.~Qi, ``Codet: Co-occurrence guided region-word alignment for open-vocabulary object detection,'' in \emph{Advances in Neural Information Processing Systems}, 2024.

\bibitem{zhao2022exploiting}
S.~Zhao, Z.~Zhang, S.~Schulter, L.~Zhao, B.~Vijay~Kumar, A.~Stathopoulos, M.~Chandraker, and D.~N. Metaxas, ``Exploiting unlabeled data with vision and language models for object detection,'' in \emph{European conference on computer vision}.\hskip 1em plus 0.5em minus 0.4em\relax Springer, 2022, pp. 159--175.

\bibitem{pham2024lp}
C.~Pham, T.~Vu, and K.~Nguyen, ``Lp-ovod: Open-vocabulary object detection by linear probing,'' in \emph{Proceedings of the IEEE/CVF Winter Conference on Applications of Computer Vision}, 2024, pp. 779--788.

\bibitem{huynh2022open}
D.~Huynh, J.~Kuen, Z.~Lin, J.~Gu, and E.~Elhamifar, ``Open-vocabulary instance segmentation via robust cross-modal pseudo-labeling,'' in \emph{Proceedings of the IEEE/CVF Conference on Computer Vision and Pattern Recognition}, 2022, pp. 7020--7031.

\bibitem{zhao2024taming}
S.~Zhao, S.~Schulter, L.~Zhao, Z.~Zhang, Y.~Suh, M.~Chandraker, D.~N. Metaxas \emph{et~al.}, ``Taming self-training for open-vocabulary object detection,'' in \emph{Proceedings of the IEEE/CVF Conference on Computer Vision and Pattern Recognition}, 2024, pp. 13\,938--13\,947.

\bibitem{xu2023dstdet}
S.~Xu, X.~Li, S.~Wu, W.~Zhang, Y.~Li, G.~Cheng, Y.~Tong, K.~Chen, and C.~C. Loy, ``Dst-det: Simple dynamic self-training for open-vocabulary object detection,'' \emph{arXiv preprint arXiv:2310.01393}, 2023.

\bibitem{detpro}
Y.~Du, F.~Wei, Z.~Zhang, M.~Shi, Y.~Gao, and G.~Li, ``Learning to prompt for open-vocabulary object detection with vision-language model,'' in \emph{Proceedings of the IEEE/CVF Conference on Computer Vision and Pattern Recognition}, 2022, pp. 14\,084--14\,093.

\bibitem{ov-detr}
Y.~Zang, W.~Li, K.~Zhou, C.~Huang, and C.~C. Loy, ``Open-vocabulary detr with conditional matching,'' in \emph{European Conference on Computer Vision}, 2022, pp. 106--122.

\bibitem{kim2024retrieval}
J.~Kim, E.~Cho, S.~Kim, and H.~J. Kim, ``Retrieval-augmented open-vocabulary object detection,'' in \emph{Proceedings of the IEEE/CVF Conference on Computer Vision and Pattern Recognition}, 2024, pp. 17\,427--17\,436.

\bibitem{zhao2024scene}
X.~Zhao, X.~Liu, D.~Wang, Y.~Gao, and Z.~Liu, ``Scene-adaptive and region-aware multi-modal prompt for open vocabulary object detection,'' in \emph{Proceedings of the IEEE/CVF Conference on Computer Vision and Pattern Recognition}, 2024, pp. 16\,741--16\,750.

\bibitem{glip}
L.~H. Li, P.~Zhang, H.~Zhang, J.~Yang, C.~Li, Y.~Zhong, L.~Wang, L.~Yuan, L.~Zhang, J.-N. Hwang \emph{et~al.}, ``Grounded language-image pre-training,'' in \emph{Proceedings of the IEEE/CVF Conference on Computer Vision and Pattern Recognition}, 2022, pp. 10\,965--10\,975.

\bibitem{grounding_dino}
S.~Liu, Z.~Zeng, T.~Ren, F.~Li, H.~Zhang, J.~Yang, C.~Li, J.~Yang, H.~Su, J.~Zhu \emph{et~al.}, ``Grounding dino: Marrying dino with grounded pre-training for open-set object detection,'' \emph{arXiv preprint arXiv:2303.05499}, 2023.

\bibitem{10480273}
H.~Shi, M.~Hayat, and J.~Cai, ``Unified open-vocabulary dense visual prediction,'' \emph{IEEE Transactions on Multimedia}, pp. 1--13, 2024.

\bibitem{fu2025llmdet}
S.~Fu, Q.~Yang, Q.~Mo, J.~Yan, X.~Wei, J.~Meng, X.~Xie, and W.-S. Zheng, ``Llmdet: Learning strong open-vocabulary object detectors under the supervision of large language models,'' \emph{arXiv preprint arXiv:2501.18954}, 2025.

\bibitem{moco}
K.~He, H.~Fan, Y.~Wu, S.~Xie, and R.~Girshick, ``Momentum contrast for unsupervised visual representation learning,'' in \emph{Proceedings of the IEEE/CVF Conference on Computer Vision and Pattern Recognition}, 2020, pp. 9729--9738.

\bibitem{nltk}
S.~Bird, E.~Klein, and E.~Loper, \emph{Natural language processing with Python: analyzing text with the natural language toolkit}.\hskip 1em plus 0.5em minus 0.4em\relax " O'Reilly Media, Inc.", 2009.

\bibitem{fang2023simple}
R.~Fang, G.~Pang, and X.~Bai, ``Simple image-level classification improves open-vocabulary object detection,'' in \emph{Proceedings of the AAAI Conference on Artificial Intelligence}, 2024, pp. 1716--1725.

\bibitem{fvlm}
W.~Kuo, Y.~Cui, X.~Gu, A.~Piergiovanni, and A.~Angelova, ``F-vlm: Open-vocabulary object detection upon frozen vision and language models,'' in \emph{International Conference on Learning Representations}, 2023.

\bibitem{wang2023learning}
T.~Wang, ``Learning to detect and segment for open vocabulary object detection,'' in \emph{Proceedings of the IEEE/CVF Conference on Computer Vision and Pattern Recognition}, 2023, pp. 7051--7060.

\bibitem{jeong2024proxydet}
J.~Jeong, G.~Park, J.~Yoo, H.~Jung, and H.~Kim, ``Proxydet: Synthesizing proxy novel classes via classwise mixup for open-vocabulary object detection,'' in \emph{Proceedings of the AAAI Conference on Artificial Intelligence}, 2024, pp. 2462--2470.

\bibitem{jin2024llms}
S.~Jin, X.~Jiang, J.~Huang, L.~Lu, and S.~Lu, ``Llms meet vlms: Boost open vocabulary object detection with fine-grained descriptors,'' \emph{ICLR}, 2024.

\bibitem{object-centric-ovd}
H.~Bangalath, M.~Maaz, M.~U. Khattak, S.~H. Khan, and F.~Shahbaz~Khan, ``Bridging the gap between object and image-level representations for open-vocabulary detection,'' in \emph{Advances in Neural Information Processing Systems}, 2022, pp. 33\,781--33\,794.

\bibitem{wu2023cora}
X.~Wu, F.~Zhu, R.~Zhao, and H.~Li, ``Cora: Adapting clip for open-vocabulary detection with region prompting and anchor pre-matching,'' in \emph{Proceedings of the IEEE/CVF Conference on Computer Vision and Pattern Recognition}, 2023, pp. 7031--7040.

\bibitem{zhong2022regionclip}
Y.~Zhong, J.~Yang, P.~Zhang, C.~Li, N.~Codella, L.~H. Li, L.~Zhou, X.~Dai, L.~Yuan, Y.~Li \emph{et~al.}, ``Regionclip: Region-based language-image pretraining,'' in \emph{Proceedings of the IEEE/CVF Conference on Computer Vision and Pattern Recognition}, 2022, pp. 16\,793--16\,803.

\bibitem{zhang2024exploring}
H.~Zhang, Q.~Zhao, L.~Zheng, H.~Zeng, Z.~Ge, T.~Li, and S.~Xu, ``Exploring region-word alignment in built-in detector for open-vocabulary object detection,'' in \emph{Proceedings of the IEEE/CVF Conference on Computer Vision and Pattern Recognition}, 2024, pp. 16\,975--16\,984.

\bibitem{wu2023clipself}
S.~Wu, W.~Zhang, L.~Xu, S.~Jin, X.~Li, W.~Liu, and C.~C. Loy, ``Clipself: Vision transformer distills itself for open-vocabulary dense prediction,'' \emph{ICLR}, 2024.

\bibitem{deng2009imagenet}
J.~Deng, W.~Dong, R.~Socher, L.-J. Li, K.~Li, and L.~Fei-Fei, ``Imagenet: A large-scale hierarchical image database,'' in \emph{Proceedings of the IEEE/CVF Conference on Computer Vision and Pattern Recognition}, 2009, pp. 248--255.

\bibitem{schuhmann2021laion}
C.~Schuhmann, R.~Vencu, R.~Beaumont, R.~Kaczmarczyk, C.~Mullis, A.~Katta, T.~Coombes, J.~Jitsev, and A.~Komatsuzaki, ``Laion-400m: Open dataset of clip-filtered 400 million image-text pairs,'' \emph{arXiv preprint arXiv:2111.02114}, 2021.

\bibitem{cc3m}
P.~Sharma, N.~Ding, S.~Goodman, and R.~Soricut, ``Conceptual captions: A cleaned, hypernymed, image alt-text dataset for automatic image captioning,'' in \emph{Proceedings of the Annual Meeting of the Association for Computational Linguistics}, 2018, pp. 2556--2565.

\bibitem{gupta2019lvis}
A.~Gupta, P.~Dollar, and R.~Girshick, ``Lvis: A dataset for large vocabulary instance segmentation,'' in \emph{Proceedings of the IEEE/CVF Conference on Computer Vision and Pattern Recognition}, 2019.

\bibitem{coco}
T.-Y. Lin, M.~Maire, S.~Belongie, J.~Hays, P.~Perona, D.~Ramanan, P.~Doll{\'a}r, and C.~L. Zitnick, ``Microsoft coco: Common objects in context,'' in \emph{European Conference on Computer Vision}, 2014.

\bibitem{shao2019objects365}
S.~Shao, Z.~Li, T.~Zhang, C.~Peng, G.~Yu, X.~Zhang, J.~Li, and J.~Sun, ``Objects365: A large-scale, high-quality dataset for object detection,'' in \emph{Proceedings of the IEEE/CVF International Conference on Computer Vision}, 2019, pp. 8430--8439.

\bibitem{resnet}
K.~He, X.~Zhang, S.~Ren, and J.~Sun, ``Deep residual learning for image recognition,'' in \emph{Proceedings of the IEEE/CVF Conference on Computer Vision and Pattern Recognition}, 2016, pp. 770--778.

\bibitem{liu2021swin}
Z.~Liu, Y.~Lin, Y.~Cao, H.~Hu, Y.~Wei, Z.~Zhang, S.~Lin, and B.~Guo, ``Swin transformer: Hierarchical vision transformer using shifted windows,'' in \emph{Proceedings of the IEEE/CVF International Conference on Computer Vision}, 2021, pp. 10\,012--10\,022.

\bibitem{mavl}
M.~Maaz, H.~Rasheed, S.~Khan, F.~S. Khan, R.~M. Anwer, and M.-H. Yang, ``Class-agnostic object detection with multi-modal transformer,'' in \emph{European Conference on Computer Vision}, 2022, pp. 512--531.

\bibitem{mokady2021clipcap}
R.~Mokady, A.~Hertz, and A.~H. Bermano, ``Clipcap: Clip prefix for image captioning,'' \emph{arXiv preprint arXiv:2111.09734}, 2021.

\bibitem{li2023blip}
J.~Li, D.~Li, S.~Savarese, and S.~Hoi, ``Blip-2: Bootstrapping language-image pre-training with frozen image encoders and large language models,'' in \emph{International Conference on Machine Learning}, 2023, pp. 19\,730--19\,742.

\bibitem{gpt2}
A.~Radford, J.~Wu, R.~Child, D.~Luan, D.~Amodei, I.~Sutskever \emph{et~al.}, ``Language models are unsupervised multitask learners,'' \emph{OpenAI blog}, vol.~1, no.~8, p.~9, 2019.

\bibitem{Flant5}
H.~W. Chung, L.~Hou, S.~Longpre, B.~Zoph, Y.~Tay, W.~Fedus, Y.~Li, X.~Wang, M.~Dehghani, S.~Brahma \emph{et~al.}, ``Scaling instruction-finetuned language models,'' \emph{Journal of Machine Learning Research}, vol.~25, no.~70, pp. 1--53, 2024.

\end{thebibliography}
%

 





\end{document}